\documentclass[11pt]{article}

\usepackage{a4wide}
\usepackage{algorithm}
\usepackage{algorithmic}
\usepackage{epsfig}
\usepackage{amsmath}
\usepackage{amsfonts}
\usepackage{amssymb}
\usepackage{pstricks}
\usepackage{pst-node}
\usepackage{subfigure}
\usepackage{multirow}
\usepackage{stmaryrd}

\usepackage[applemac]{inputenc}
\usepackage[german,spanish,catalan,english]{babel}

\newcommand{\fin}{$\text{Finocchi}$}
\newcommand{\fsim}{$\text{FrogSim}^{\varominus}$}
\newcommand{\fsimp}{$\text{FrogSim}$}

\begin{document}

\title{Distributed Graph Coloring: An Approach Based on the Calling Behavior of Japanese Tree Frogs}

\author{H.~Hern{\'a}ndez and C.~Blum \\
~\\
ALBCOM Research Group\\ 
Universitat Polit{\`e}cnica de Catalunya, Barcelona, Spain \\
{\sf \{hhernandez,cblum\}@lsi.upc.edu}}

\date{}

% make the title area
\maketitle

\begin{abstract}
Graph coloring---also known as vertex coloring---considers the problem of assigning colors to the nodes of a graph such that adjacent nodes do not share the same color. The optimization version of the problem concerns the minimization of the number of used colors. In this paper we deal with the problem of finding valid colorings of graphs in a distributed way, that is, by means of an algorithm that only uses local information for deciding the color of the nodes. Such algorithms prescind from any central control. Due to the fact that quite a few practical applications require to find colorings in a distributed way, the interest in distributed algorithms for graph coloring has been growing during the last decade. As an example consider wireless ad-hoc and sensor networks, where tasks such as the assignment of frequencies or the assignment of TDMA slots are strongly related to graph coloring. 

The algorithm proposed in this paper is inspired by the calling behavior of Japanese tree frogs. Male frogs use their calls to attract females. Interestingly, groups of males that are located nearby each other desynchronize their calls. This is because female frogs are only able to correctly localize the male frogs when their calls are not too close in time. We experimentally show that our algorithm is very competitive with the current state of the art, using different sets of problem instances and comparing to one of the most competitive algorithms from the literature.
\end{abstract}

% The body of the paper starts here:
\section{Introduction}

Given an undirected graph $G=(V,E)$, where $V$ is the node set and $E$ is the edge set, and a number $k > 0$ of colors, a valid \emph{$k$-coloring} of the graph is the assignment of exactly one color to each node such that adjacent nodes (that is, nodes that are connected by an edge) do not share the same color. Formally, we say that a $k$-coloring of an undirected graph $G=(V,E)$ is a function $c:V\rightarrow \{1,2,\ldots,k\}$ such that $c(u) \neq c(v)$ for each edge $(u,v)\in E$. The optimization version of the \emph{graph coloring problem} (GCP), which is $NP$-hard~\cite{karp1972reducibility}, consists in finding the minimum number $k^*$ of colors such that a valid $k^*$-coloring can be found. This number is called the chromatic number of graph $G$ and is denoted by $\chi(G)$. The GCP is a quite generic problem. Practical applications originate especially from problems that can be modelled by networks and graphs, for example, communication networks. Several tasks in modern wireless ad-hoc networks, such as sensor networks, are related to graph coloring. Examples include TDMA slot assignment~\cite{herman2004distributed}, detection of mobile objects and reduction of signaling actuators~\cite{zhang2005distributed}, distributed MAC layer management~\cite{guo2001low}, energy-efficient coverage~\cite{cardei2002wireless}, delay efficient sleep scheduling~\cite{lu2005delay} or wakeup scheduling~\cite{keshavarzian2006wakeup}. Due to the distributed nature of these networks, algorithms for solving problems related to graph coloring are generally also required to be distributed~\cite{Luy09:book}. Such algorithms make an exclusive use of local information for deciding the color of the nodes, that is, they are characterized by the absence of any central control mechanism. The goal of this paper is to device an algorithm for generating valid colorings in a distributed manner.  \\ 

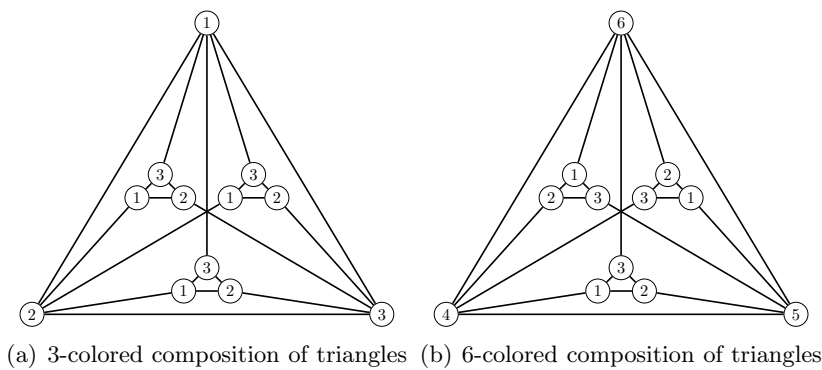
\begin{figure}[ht!]
\centering
\subfigure[3-colored composition of triangles]{
\label{fig:simple-compotriangle-correct}
\scalebox{0.62}{
\psset{xunit=0.5mm,yunit=0.5mm,runit=0.5mm}
\begin{pspicture}(-30,-5)(130,130)

\rput(40,10){\circlenode[linewidth=0.5pt,fillstyle=solid]{1}{\small 1}}
\rput(60,10){\circlenode[linewidth=0.5pt,fillstyle=solid]{2}{\small 2}}
\rput(50,20){\circlenode[linewidth=0.5pt,fillstyle=solid]{3}{\small 3}}

\rput(20,50){\circlenode[linewidth=0.5pt,fillstyle=solid]{4}{\small 1}}
\rput(40,50){\circlenode[linewidth=0.5pt,fillstyle=solid]{5}{\small 2}}
\rput(30,60){\circlenode[linewidth=0.5pt,fillstyle=solid]{6}{\small 3}}

\rput(60,50){\circlenode[linewidth=0.5pt,fillstyle=solid]{7}{\small 1}}
\rput(80,50){\circlenode[linewidth=0.5pt,fillstyle=solid]{8}{\small 2}}
\rput(70,60){\circlenode[linewidth=0.5pt,fillstyle=solid]{9}{\small 3}}

\rput(-25,0){\circlenode[linewidth=0.5pt,fillstyle=solid]{10}{\small 2}}
\rput(125,0){\circlenode[linewidth=0.5pt,fillstyle=solid]{11}{\small 3}}
\rput(50,125){\circlenode[linewidth=0.5pt,fillstyle=solid]{12}{\small 1}}

\ncline[linewidth=1pt]{-}{1}{2}
\ncline[linewidth=1pt]{-}{2}{3}
\ncline[linewidth=1pt]{-}{3}{1}

\ncline[linewidth=1pt]{-}{4}{5}
\ncline[linewidth=1pt]{-}{5}{6}
\ncline[linewidth=1pt]{-}{6}{4}

\ncline[linewidth=1pt]{-}{7}{8}
\ncline[linewidth=1pt]{-}{8}{9}
\ncline[linewidth=1pt]{-}{9}{7}

\ncline[linewidth=1pt]{-}{10}{1}
\ncline[linewidth=1pt]{-}{10}{4}
\ncline[linewidth=1pt]{-}{10}{7}

\ncline[linewidth=1pt]{-}{11}{2}
\ncline[linewidth=1pt]{-}{11}{5}
\ncline[linewidth=1pt]{-}{11}{8}

\ncline[linewidth=1pt]{-}{12}{3}
\ncline[linewidth=1pt]{-}{12}{6}
\ncline[linewidth=1pt]{-}{12}{9}

\ncline[linewidth=1pt]{-}{10}{11}
\ncline[linewidth=1pt]{-}{11}{12}
\ncline[linewidth=1pt]{-}{12}{10}
\end{pspicture}
}
}
\subfigure[6-colored composition of triangles]{
\label{fig:simple-compotriangle-bad}
\scalebox{0.62}{
\psset{xunit=0.5mm,yunit=0.5mm,runit=0.5mm}
\begin{pspicture}(-30,-5)(130,130)

\rput(40,10){\circlenode[linewidth=0.5pt,fillstyle=solid]{1}{\small 1}}
\rput(60,10){\circlenode[linewidth=0.5pt,fillstyle=solid]{2}{\small 2}}
\rput(50,20){\circlenode[linewidth=0.5pt,fillstyle=solid]{3}{\small 3}}

\rput(20,50){\circlenode[linewidth=0.5pt,fillstyle=solid]{4}{\small 2}}
\rput(40,50){\circlenode[linewidth=0.5pt,fillstyle=solid]{5}{\small 3}}
\rput(30,60){\circlenode[linewidth=0.5pt,fillstyle=solid]{6}{\small 1}}

\rput(60,50){\circlenode[linewidth=0.5pt,fillstyle=solid]{7}{\small 3}}
\rput(80,50){\circlenode[linewidth=0.5pt,fillstyle=solid]{8}{\small 1}}
\rput(70,60){\circlenode[linewidth=0.5pt,fillstyle=solid]{9}{\small 2}}

\rput(-25,0){\circlenode[linewidth=0.5pt,fillstyle=solid]{10}{\small 4}}
\rput(125,0){\circlenode[linewidth=0.5pt,fillstyle=solid]{11}{\small 5}}
\rput(50,125){\circlenode[linewidth=0.5pt,fillstyle=solid]{12}{\small 6}}

\ncline[linewidth=1pt]{-}{1}{2}
\ncline[linewidth=1pt]{-}{2}{3}
\ncline[linewidth=1pt]{-}{3}{1}

\ncline[linewidth=1pt]{-}{4}{5}
\ncline[linewidth=1pt]{-}{5}{6}
\ncline[linewidth=1pt]{-}{6}{4}

\ncline[linewidth=1pt]{-}{7}{8}
\ncline[linewidth=1pt]{-}{8}{9}
\ncline[linewidth=1pt]{-}{9}{7}

\ncline[linewidth=1pt]{-}{10}{1}
\ncline[linewidth=1pt]{-}{10}{4}
\ncline[linewidth=1pt]{-}{10}{7}

\ncline[linewidth=1pt]{-}{11}{2}
\ncline[linewidth=1pt]{-}{11}{5}
\ncline[linewidth=1pt]{-}{11}{8}

\ncline[linewidth=1pt]{-}{12}{3}
\ncline[linewidth=1pt]{-}{12}{6}
\ncline[linewidth=1pt]{-}{12}{9}

\ncline[linewidth=1pt]{-}{10}{11}
\ncline[linewidth=1pt]{-}{11}{12}
\ncline[linewidth=1pt]{-}{12}{10}

\end{pspicture}
}
}
\caption{Simple graph topology (composed of three inner triangles and one outer triangle). (a) shows an optimal 3-coloring, while (b) shows a sub-optimal 6-coloring. Distributed algorithms provide most often a 6-colored solution, because global knowledge is necessary for capturing the graph structure.}
\label{fig:simple}
\end{figure}

The distributed conception of an algorithm is generally beneficial for its scalability. Moreover, in comparison to centralized approaches it is generally much easier to adapt a distributed algorithm to dynamic changes during execution. Unfortunately, the exclusive use of local information is often not sufficient to completely capture the internal structure of certain graphs or networks. The following example helps to understand the tradeoff between generating colorings from a local and a global perspective. Figure~\ref{fig:simple} shows a graph which has been constructed using four different triangles, that is, complete graphs of three nodes. Hereby, we distinguish between three \emph{inner triangles} (the three groups of nodes that are close together) and one \emph{outer triangle}. The three inner triangles are connected to the outer triangle such that each node of a specific inner triangle is connected to a different node of the outer triangle. Even in a distributed manner it is fairly easy to obtain optimal colorings for each of the inner triangles. Depending on the specific color assignment concerning the three inner triangles the outer triangle may be colored with the same three colors (as in Figure~\ref{fig:simple-compotriangle-correct}) or with three additional colors (as in Figure~\ref{fig:simple-compotriangle-bad}). Unfortunately, probability for the latter case is quite high, especially when the complexity of the graph is increased by adding more inner triangles. As mentioned already above, one of the key difficulties when coloring graphs in a distributed manner is that each node is only provided with local information and, therefore, it is unable to detect situations such as the one from Figure~\ref{fig:simple-compotriangle-bad}. \\

\subsection{Our Contribution}

In this paper we propose a distributed algorithm for graph coloring based on the calling behavior exhibited by male \emph{Japanese tree frogs} for the attraction of females. Several researchers have observed that male \emph{Japanese tree frogs} decouple their calls~\cite{wells1977social}. This property has evolved because females can only localize the males when their calling is not too close in time. In~\cite{aihara2008mathematical} Aihara et al.~proposed a theoretical model for simulating the behavior of these frogs. The authors describe an oscillator system, where each oscillator has a phase $\theta \in [0,2\pi] $ that changes over time with frequency $\omega$ (where $2\pi$ is the time interval between two calls of the same frog). When the phase reaches $2\pi$, the oscillator fires and returns to the baseline phase ($\theta=0$). The proposed system works such that oscillators try to maximize the distance between their phases. This model works nicely for the desynchronization of two oscillators. However, when more than two oscillators are concerned, the model does not accurately reflect the real behavior of the frogs. A subsequent work~\cite{aihara2009:modeling} mentions some potential applications of this model in artificial life and robotics. In both works the author(s) mention the limitations of the systems when operating with groups of more than two coupled oscillators. In fact, already with three oscillators the final solution (and its stability) strongly depends on the initial variable settings. 

The desynchronization of the frogs' calls is achieved in a self-organized way. Therefore, the algorithm proposed in this paper, which is based on this self-desynchronization mechanism, can be regarded as a \emph{swarm intelligence} approach~\cite{BonDorThe99:book,BluMer08:book}. Swarm intelligence is a field of computer science which is inspired by the collective behavior of social animals and other self-organizing processes from nature. Successful examples from the literature include \emph{particle swarm optimization} (PSO)~\cite{kennedy1995particle}, which is an algorithm for optimization inspired by bird flocking and fish schooling, and \emph{ant colony optimization} (ACO)~\cite{DorStu04:book}, which is inspired by the foraging behavior of ant colonies. One of the distinguishing properties of a swarm intelligence approach is the fact that the problem at hand is solved from a local perspective. Moreover, problem solving is based on the cooperation of rather simple entities. Instead of each entity trying to solve the problem by itself, they perform simple tasks from a local perspective. The global problem is solved as a result of cooperation. Therefore, swarm intelligence principles are well suited for their use in distributed algorithms.

The proposed algorithm uses a desynchronization method based on the original model by Aihara et al.~\cite{aihara2008mathematical}, with some small modifications. The algorithm can be easily implemented, for example, in sensor networks. In addition to competitive results it comes with several advantages as, for example, a low consumption of energy resources or its potential ability to adapt to changes in the network topology. However, as mentioned before, the main goal of the algorithm is to obtain valid colorings that use an as-low-as-possible number of colors, while keeping the number of iterations necessary to reach these results as low as possible. An extensive experimental evaluation shows that the results of the algorithm are comparable or better than the ones of state-of-the-art algorithms for what concerns the number colors. In particular, the good performance of our algorithm for grid graphs of any size is remarkable. On the downside, the results also show that our algorithm may require a slightly higher number of communication rounds than other state-of-the-art algorithms.

\subsection{Prior Work on Graph Coloring}

Concerning prior work, a distinction must be made between centralized and distributed algorithms. Concerning centralized algorithms, the literature offers both exact approaches that guarantee to find an optimal solution in bounded time and (meta-)heuristic approaches. A recent survey can be found in~\cite{malaguti2010survey}. Due to the intractable nature of the GCP, larger problem instances can only be tackled efficiently by heuristic approaches. Especially effective are the tabu search algorithm from~\cite{blochliger2008graph}, a hybrid approach combining tabu search and evolutionary algorithms from~\cite{lu2010memetic} and a variable neighborhood search technique~\cite{hertz2008variable}. These algorithms are nowadays the best centralized metaheuristics for solving the GCP. 

When considering distributed algorithms, it is very difficult (if not impossible) to narrow down the state of the art to a small set of algorithms. This is because distributed algorithms may be designed with very different goals. These goals may concern, for example, the performance for particular topologies, the minimization of execution time (or communication rounds), the generation of the best colorings possible, or the performance for dynamically changing topologies. In addition, a general problem is that most proposals are not evaluated on publicly available sets of benchmark instances. Moreover, results are generally not shown per instance, making it difficult to compare to the proposed algorithms. In the following we only focus on algorithms that generate valid solutions and possibly refer to their simplicity, solution quality and time complexity.\footnote{In the scope of this paper the time complexity is, as usual, measured in terms of \emph{communication rounds}. A communication round is the unit of time in which each node is allowed to send at most one message.} It must also be noted that many of the proposed distributed algorithms were developed for applications in networks of devices with scarce resources. For this reason authors often study the message load the algorithm implies and try to minimize the amount of calculus required by the algorithm. Typically, these algorithms are meant to work on a lower layer of the network in parallel with the applications or information flows that the user may require to send. In~\cite{fraigniaud2009distributed}, Fraigniaud et al. study the effect of the amount of information shared between the nodes on the quality of the obtained colorings.

One of the most general works was presented by Finocchi et al.~in~\cite{finocchi2005experimental}. The authors introduced three versions of a distributed algorithm and study its behavior under various conditions. The authors considered both the problem of obtaining $O(\Delta+1)$-colorings in as few communication rounds as possible, as well as the problem of generating the best possible colorings without any limit on the number of communication rounds. The authors provide extensive experimental results for both cases. Most of their experimentation is based on random graphs, which are not publicly available. However, they also offer results on a well-known set of publicly available instances from the DIMACS challenge~\cite{dimacschallenge}. As the algorithm proposed in~\cite{finocchi2005experimental} was shown to outperform the state of the art, we have chosen this algorithm for comparison.

Concerning distributed algorithms based on swarm intelligence principles, the literature offers, for example, a method inspired by the synchronous flashing of fireflies (see~\cite{lee2008firefly}). This algorithm, which allows a simple implementation, reaches valid colorings fast, in a constant number of communication rounds, regardless of the size of the network. However, this work does not focus on minimizing the number of colors. The first intent to use the calling behavior of frogs for graph coloring was presented in~\cite{leelister2008:coupled}. Valid colorings are obtained by assigning a color to each phase used by the nodes (that is, the oscillators). Therefore, if two nodes are synchronized to exactly the same phase, they will be sharing a common color (the authors consider a function $f:[0,2\pi] \rightarrow (R,G,B)$, where $2\pi$ is the time frame between two callings of the same frog). The main drawback of this approach is that nodes with very near phases will be colored with different colors. As such small deviations usually occur when the number of nodes in the system increases, the algorithm does not obtain competitive results. This work was further extended by adding a parameter for setting a priori the number of allowed phases~\cite{lee2010k}. Experimentation shows that the system is able to find optimal solutions for small topologies, provided the optimal number of colors is known. Note that in contrast to these works, the algorithm that we propose aims for the minimization of the used number of colors without any prior knowledge about the optimal solution.

The literature also offers many works that consider distributed graph coloring from a theoretical point of view. Most of them concern upper bounds for the coloring quality as well as the time complexity under different constraints. Hansen et al.~\cite{hansen2004distributed} proposed the \emph{distributed largest-first} (DLF) algorithm that runs in $O(\Delta^2 \text{log}n)$ communication rounds for arbitrary graphs and that was proven to provide good upper bounds for specific topologies. This algorithm was based on the \emph{largest-first} approach which consists in giving priority for choosing a color to the nodes with the highest degree ($\Delta$). This work was further extended by Kosowski and Kuszner~\cite{kosowski2006greedy} who reduced the time complexity to $O(\Delta \text{log}n \text{log}\Delta)$. These authors also proved that some other approaches, like \emph{smallest-last} or \emph{dynamic-saturation}, are not suitable for distributed environments. Later, in~\cite{moscibroda2008coloring} Moscibroda and Wattenh{\"o}fer introduced an algorithm for obtaining $O(\Delta)$-colorings in $O(\delta \text{log}n)$ time when considering random geometric graphs and other well-known models for wireless multi-hop networks (no results are given for other topologies). Other theoretical works which may be of interest for the development of new algorithms are the game theoretic approach for efficient graph coloring from Panagopoulou and Spirakis~\cite{panagopoulou2008game} and the work by Kuhn and Wattenh{\"o}fer~\cite{kuhn2006complexity}, which introduces a new lower bound on the number of colors used by algorithms that are restricted to one single communication round and a new lower bound on the time complexity of obtaining a $O(\Delta)$-coloring of a graph.

\subsection{Organization of the Paper}

The rest of this paper is organized as follows. Section~\ref{sec:frogs-in-nature} describes the behavior of frogs in nature, which has inspired our algorithm. Moreover, existing models are outlined. In Section~\ref{sec:algorithm} the algorithm is introduced. An extensive experimental evaluation of the proposed algorithm is presented in Section~\ref{sec:results}. Finally, Section~\ref{sec:conclusions} is dedicated to conclusions and the outline of future work.

\section{Modelling the Calling Behavior of Japanese Tree Frogs}
\label{sec:frogs-in-nature}

Different studies (see, for example,~\cite{wells1977social}) have shown that male Japanese tree frogs use their calling to attract females. Apparently, females of this family of frogs can recognize the source of the calling in order to determine the current location of the corresponding male. A problem arises when two of these males are too close in space and communicate at the same time. In this case females are not able to properly recognize both calls independently and are, therefore, unable to detect where the calls came from. For this reason, males have evolved to desynchronize their sounds in time. They achieve to uniformly distribute the distance between each pair of calls, which allows the females to locate the males they can hear, and to choose one. In fact, this behavior is a prime example for self-organization in nature.

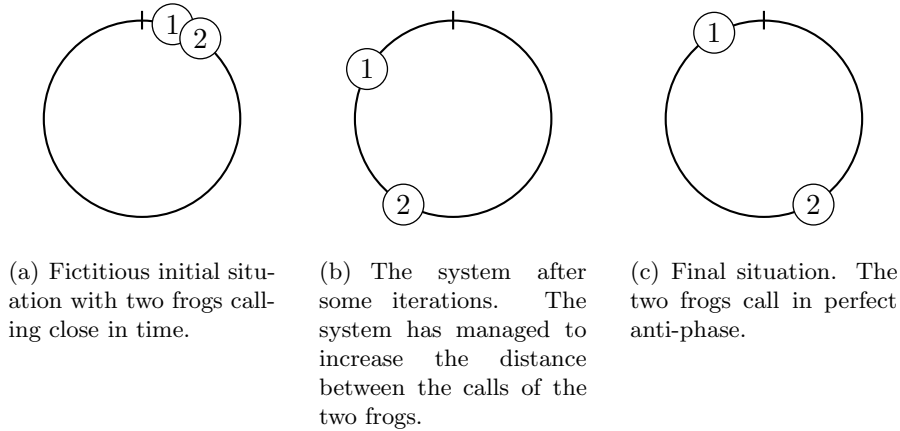
\begin{figure}[!t]
\centering
\subfigure[Fictitious initial situation with two frogs calling close in time.]{
\label{fig:sample-behavior-1}
\psset{xunit=.33mm,yunit=.33mm,runit=.33mm}
\begin{pspicture}(100,100)
\pscircle(50,50){40}
\psline(50,93.5)(50,86)
\rput(62.36,88.04){\circlenode[linewidth=0.5pt,fillstyle=solid,fillcolor=white]{1}{$1$}}
\rput(73.51,82.36){\circlenode[linewidth=0.5pt,fillstyle=solid,fillcolor=white]{2}{$2$}}
 \end{pspicture}
}
$\;\;\;$
\subfigure[The system after some iterations. The system has managed to increase the distance between the calls of the two frogs.]{
\label{fig:sample-behavior-2}
\psset{xunit=.33mm,yunit=.33mm,runit=.33mm}
\begin{pspicture}(100,100)
\pscircle(50,50){40}
\psline(50,93.5)(50,86)
\rput(15.36,70.00){\circlenode[linewidth=0.5pt,fillstyle=solid,fillcolor=white]{1}{$1$}}
\rput(30.00,15.36){\circlenode[linewidth=0.5pt,fillstyle=solid,fillcolor=white]{2}{$2$}}
 \end{pspicture}
}
$\;\;\;$
\subfigure[Final situation. The two frogs call in perfect anti-phase.]{
\label{fig:sample-behavior-3}
\psset{xunit=.33mm,yunit=.33mm,runit=.33mm}
\begin{pspicture}(100,100)
\pscircle(50,50){40}
\psline(50,93.5)(50,86)
\rput(30,84.64){\circlenode[linewidth=0.5pt,fillstyle=solid,fillcolor=white]{1}{$1$}}
\rput(70.00,15.36){\circlenode[linewidth=0.5pt,fillstyle=solid,fillcolor=white]{2}{$2$}}
 \end{pspicture}
}
\caption{Graphical illustration of the working of a system of two coupled oscillators. The circle in all three graphics represents the time frame between two calls of the same frog ($2\pi$), the calling period. The nodes marked by integer numbers  1 and 2 indicate the phase of the corresponding frogs, that is, the moment of time in which they call. (a) shows a fictitious initial situation. (b) shows the situation after some iterations. Clearly the system tries to put some distance between the calling of frogs 1 and 2. (c) shows an optimal final situation in which the frogs (or oscillators) are in perfect anti-phase, that is, their respective calls have the reached the maximum distance in time (half a circle).}
\label{fig:sample-behavior}
\end{figure}

More recently, Aihara et al.~\cite{aihara2008mathematical} introduced a formal model based on a set of coupled oscillators each one simulating the phase change in the calling period of a single frog. As oscillators are associated to frogs, we will use both terms in the following with the same meaning. The basic way of working of this model is graphically illustrated in Figure~\ref{fig:sample-behavior}. The circle represents---in all three graphics---the time frame between two calls of the same frog ($2\pi$), the calling period. The nodes marked by integer numbers 1 and 2 indicate the phase of the corresponding frogs, that is, the moment of time in which they call. Note that the oscillators are not able to reach perfect anti-phase in a single step. In general, an indefinite number of steps is needed before reaching the stable situation corresponding to perfect anti-phase. Moreover, the difficulty of reaching the optimal configuration tends to increase with an increasing number of frogs and also with an increasing degree of interaction between them (note that two frogs that can not hear each other do not influence each other). \\

\begin{figure}[!t]
\centering
\subfigure[Topology 1]{
\label{fig:bad-example-t1}
\psset{xunit=.33mm,yunit=.33mm,runit=.33mm}
\begin{pspicture}(100,100)
\rput(20,20){\circlenode[linewidth=0.5pt,fillstyle=solid,fillcolor=white]{1}{$1$}}
\rput(80,20){\circlenode[linewidth=0.5pt,fillstyle=solid,fillcolor=white]{2}{$2$}}
\rput(80,80){\circlenode[linewidth=0.5pt,fillstyle=solid,fillcolor=white]{3}{$3$}}
\rput(20,80){\circlenode[linewidth=0.5pt,fillstyle=solid,fillcolor=white]{4}{$4$}}

\ncline[linewidth=0.5pt]{-}{1}{2}
\ncline[linewidth=0.5pt]{-}{2}{3}
\ncline[linewidth=0.5pt]{-}{3}{4}
\ncline[linewidth=0.5pt]{-}{4}{1}
 \end{pspicture}
}
$\;\;\;$
\subfigure[Suboptimal desynchronization of topology 1 (with 4 different phases)]{
\label{fig:bad-example-t1-bad}
\psset{xunit=.33mm,yunit=.33mm,runit=.33mm}
\begin{pspicture}(100,100)
\pscircle(50,50){40}
\psline(50,93.5)(50,86)
\rput(78.28,78.28){\circlenode[linewidth=0.5pt,fillstyle=solid,fillcolor=white]{1}{$1$}}
\rput(78.28,21.72){\circlenode[linewidth=0.5pt,fillstyle=solid,fillcolor=white]{2}{$2$}}
\rput(21.72,21.72){\circlenode[linewidth=0.5pt,fillstyle=solid,fillcolor=white]{3}{$3$}}
\rput(21.72,78.28){\circlenode[linewidth=0.5pt,fillstyle=solid,fillcolor=white]{4}{$4$}}
\ncline[linewidth=0.5pt]{-}{1}{2}
\ncline[linewidth=0.5pt]{-}{2}{3}
\ncline[linewidth=0.5pt]{-}{3}{4}
\ncline[linewidth=0.5pt]{-}{4}{1}
\end{pspicture}
}
$\;\;\;$
\subfigure[Optimal desynchronization of topology 1 (with 2 different phases)]{
\label{fig:bad-example-t1-optimal}
\psset{xunit=.33mm,yunit=.33mm,runit=.33mm}
\begin{pspicture}(100,100)
\pscircle(50,50){40}
\psline(50,93.5)(50,86)
\rput(78.28,78.28){\circlenode[linewidth=0.5pt,fillstyle=solid,fillcolor=white]{1}{$1,3$}}
%\rput(78.28,21.72){\circlenode[linewidth=0.5pt,fillstyle=solid,fillcolor=white]{2}{$2$}}
\rput(21.72,21.72){\circlenode[linewidth=0.5pt,fillstyle=solid,fillcolor=white]{2}{$2,4$}}
%\rput(21.72,78.28){\circlenode[linewidth=0.5pt,fillstyle=solid,fillcolor=white]{4}{$4$}}
\ncline[linewidth=0.5pt]{-}{1}{2}
%\ncline[linewidth=0.5pt]{-}{2}{3}
%\ncline[linewidth=0.5pt]{-}{3}{4}
%\ncline[linewidth=0.5pt]{-}{4}{1}
 \end{pspicture}
}
\\
\subfigure[Topology 2]{
\label{fig:bad-example-t2}
\psset{xunit=.33mm,yunit=.33mm,runit=.33mm}
\begin{pspicture}(100,100)
\rput(20,20){\circlenode[linewidth=0.5pt,fillstyle=solid,fillcolor=white]{1}{$1$}}
\rput(80,20){\circlenode[linewidth=0.5pt,fillstyle=solid,fillcolor=white]{2}{$2$}}
\rput(80,80){\circlenode[linewidth=0.5pt,fillstyle=solid,fillcolor=white]{3}{$3$}}
\rput(20,80){\circlenode[linewidth=0.5pt,fillstyle=solid,fillcolor=white]{4}{$4$}}
\rput(50,50){\circlenode[linewidth=0.5pt,fillstyle=solid,fillcolor=white]{5}{$5$}}

\ncline[linewidth=0.5pt]{-}{1}{2}
\ncline[linewidth=0.5pt]{-}{2}{3}
\ncline[linewidth=0.5pt]{-}{3}{4}
\ncline[linewidth=0.5pt]{-}{4}{1}
\ncline[linewidth=0.5pt]{-}{1}{5}
\ncline[linewidth=0.5pt]{-}{2}{5}
\ncline[linewidth=0.5pt]{-}{3}{5}
\ncline[linewidth=0.5pt]{-}{4}{5}
\end{pspicture}
}
$\;\;\;$
\subfigure[Suboptimal desynchronization of topology 2 (with 4 different phases)]{
\label{fig:bad-example-t2-bad}
\psset{xunit=.33mm,yunit=.33mm,runit=.33mm}
\begin{pspicture}(100,100)
\pscircle(50,50){40}
\psline(50,93.5)(50,86)
\rput(78.28,78.28){\circlenode[linewidth=0.5pt,fillstyle=solid,fillcolor=white]{1}{$1$}}
\rput(78.28,21.72){\circlenode[linewidth=0.5pt,fillstyle=solid,fillcolor=white]{2}{$2,4$}}
\rput(21.72,21.72){\circlenode[linewidth=0.5pt,fillstyle=solid,fillcolor=white]{3}{$3$}}
\rput(21.72,78.28){\circlenode[linewidth=0.5pt,fillstyle=solid,fillcolor=white]{5}{$5$}}

\ncline[linewidth=0.5pt]{-}{1}{2}
\ncline[linewidth=0.5pt]{-}{1}{5}
\ncline[linewidth=0.5pt]{-}{3}{2}
\ncline[linewidth=0.5pt]{-}{3}{5}
\ncline[linewidth=0.5pt]{-}{2}{5}
\end{pspicture}
}
$\;\;\;$
\subfigure[Optimal desynchronization of topology 2  (with three different phases)]{
\label{fig:bad-example-t2-optimal}
\psset{xunit=.33mm,yunit=.33mm,runit=.33mm}
\begin{pspicture}(100,100)
\pscircle(50,50){40}
\psline(50,93.5)(50,86)
\rput(84.64,70.00){\circlenode[linewidth=0.5pt,fillstyle=solid,fillcolor=white]{1}{$1,3$}}
\rput(15.36,70.00){\circlenode[linewidth=0.5pt,fillstyle=solid,fillcolor=white]{2}{$2,4$}}
\rput(50.00,10.00){\circlenode[linewidth=0.5pt,fillstyle=solid,fillcolor=white]{5}{$5$}}

\ncline[linewidth=0.5pt]{-}{1}{2}
\ncline[linewidth=0.5pt]{-}{2}{5}
\ncline[linewidth=0.5pt]{-}{5}{1}
\end{pspicture}
}
\caption{Two examples for graph topologies (graphics (a) and (d)) that may cause problems for the desynchronization as performed by the model proposed in~\cite{aihara2008mathematical}. Graphics (b) and (e) show suboptimal desynchronizations (corresponding to stable attractors of the system) for both topologies. In contrast, graphics (c) and (f) show optimal desynchronizations.}
\label{fig:bad-example}
\end{figure}
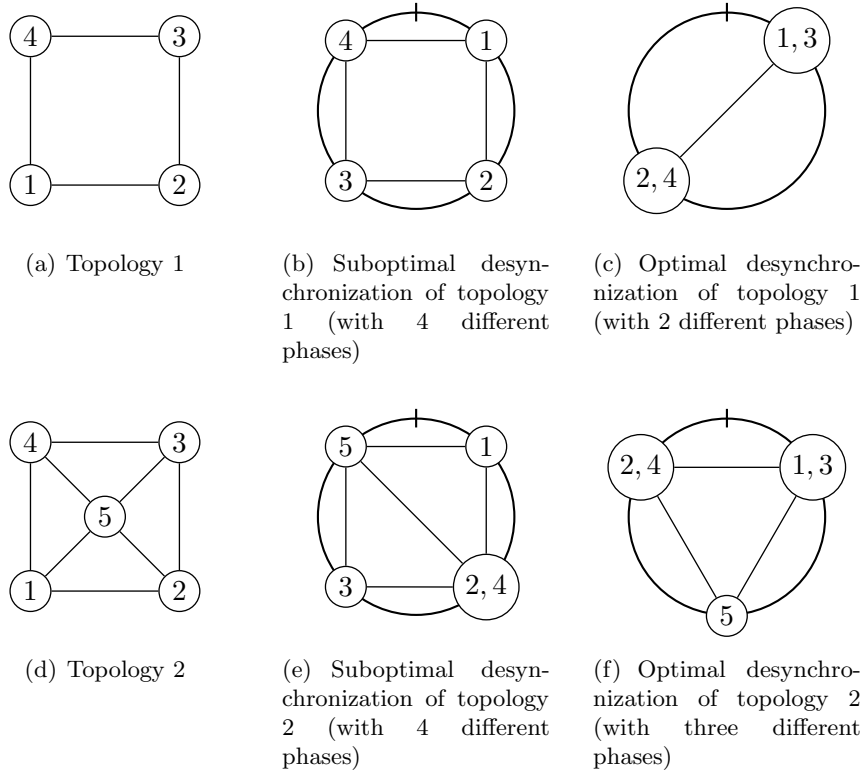

Technically, the system introduced by Aihara et al.~\cite{aihara2008mathematical} works as follows. Each oscillator $i$ has a phase $\theta_i \in [0,2 \pi] $ that changes over time with frequency $\omega_i$ (where $2\pi$ is the time interval between two calls of the same frog, the calling period). When the phase reaches $2\pi$, the oscillator fires and returns to the baseline. In addition, oscillators may be coupled with other oscillators. In case an oscillator $j$ is coupled to an oscillator $i$, when oscillator $i$ fires, oscillator $j$ receives a boost and changes the frequency of firing in the next round depending on the gap $\Delta_ {ji}\in [0,2\pi] $ (see below) between both oscillators. These changes do not happen instantly upon receiving the stimulus. The corresponding oscillator rather waits until it fires. The model can be summarized in the following equations. First, the behavior of an isolated oscillator $i$ is modelled as follows:
\begin{equation}
\frac{d\theta_i}{dt}=\omega_i
\end{equation}
Assuming that oscillators $j$ and $i$ are coupled, the gap between their (current) phases is defined as:
\begin{equation}
\Delta_{ji} = \theta_j - \theta_i
\end{equation}
Now, the change in the behavior of oscillator $j$ as influenced by oscillator $i$ can be described as follows:
\begin{equation}
\label{eq:phase-change}
\frac{d\theta_j}{dt}=\omega_j + g(\Delta_{ji}) \quad,
\end{equation}
where $g(\cdot)$ is the phase shift function which is responsible for changing the phase of the frogs that are influenced by other frogs. In~\cite{aihara2008mathematical}, the authors suggest the use of the following phase shift function:
\begin{equation}
\label{eq:phase-shift}
g(x) = \alpha \sin(x)
\end{equation}
We say that this system of oscillators is in a stable situation and in anti-phase when the following two conditions are satisfied:
\begin{eqnarray}
\Delta_{ij}    & = & \Delta_{ji} \quad, \\
g(\Delta_{ij}) & = & 0 \quad,
\end{eqnarray}
for all $i \not= j$. The system presented in~\cite{aihara2008mathematical} is able to successfully locate two coupled oscillators in perfect anti-phase, independent of the initial settings of $\theta_1$ and $\theta_2$. Unfortunately, several problems arise when the number of oscillators grows. Figure~\ref{fig:bad-example} shows two examples for such problems. Given an undirected graph $G=(V,E)$, henceforth we will assign one oscillator to each node in the graph. Therefore, in the following the terms \emph{node} and \emph{oscillator} will refer to the same. We consider that two oscillators are coupled if and only if their corresponding nodes are connected by an edge. Depending on the initial phases of the oscillators, for both topologies shown in Figures~\ref{fig:bad-example-t1} and~\ref{fig:bad-example-t2} it is possible to reach suboptimal desynchronizations (as shown in Figures~\ref{fig:bad-example-t1-bad} and~\ref{fig:bad-example-t2-bad}). The corresponding optimal desynchronizations are shown in Figures~\ref{fig:bad-example-t1-optimal} and~\ref{fig:bad-example-t2-optimal}. In~\cite{aihara2008mathematical} the authors provide analytical results for using three oscillators and show that there is a high system sensitivity with respect to the initial phases (only a small subset of the possible initial settings leads to an optimal solution). \\

The initial model by Aihara et al.~\cite{aihara2008mathematical} was later extended by Mutazono et al.~\cite{mutazono2009frog}. They used their extended model for anti-phase synchronization for the purpose of collision-free transmission scheduling in sensor networks. In order to make the system applicable to larger topologies (sensor networks may consists of hundreds of nodes), they introduced weights in order to regulate the coupling between each pair of oscillators. The resulting phase shift function as introduced in~\cite{mutazono2009frog} can be described as follows:

\begin{eqnarray}
\delta(x) & = & \text{min}\{x,2\pi-x\} \quad, \\
g(x)      & = & \alpha \text{sin}(x) \cdot e^{-\delta(x)}
\end{eqnarray}
Thanks to these weights, the system reaches stable situations more easily, especially when rather small values of $\alpha$ are used. The authors experimented with topologies of up to $20$ nodes and although the system still showed certain difficulties to reach stable solutions, the sensitivity to initial conditions decreased significantly. 

Mutazono et al.~\cite{mutazono2009frog} compared the results of their system to another mechanism for coupled oscillator desynchronization proposed in~\cite{degesys2008towards}. Note that the mechanism from~\cite{degesys2008towards} is not based on the calling behavior of Japanese tree frogs. The main difference to frog-inspired systems is the fact that the phase change of a node is made on the basis of only two other nodes. The phase values allow to order all the nodes sequentially from small to large phase values. The nodes whose phase values are used to change the phase value of a node are determined as the predecessor and the successor in this (cyclic) sequence. As shown in~\cite{mutazono2009frog}, both systems achieve similar results although no extensive experimentation is made on a broad-enough set of network topologies: mostly random geometric graphs and hand-made instances with at most eight nodes were used. 

Another extension of the system by Aihara et al.~\cite{aihara2008mathematical} was introduced in~\cite{leelister2008:coupled}. The changes concern the use of different weights for the phase shift function and the introduction of a so-called frustration parameter which reduces the coupling between each pair of nodes. The authors show that their system is able to obtain better solutions than the original model for many different topologies as, for example, $k$-partite graphs, grids or platonic solids. Moreover, the authors make some interesting observations: (1) the number of oscillators is not the key factor for achieving desynchronization. It is rather the topology which most determines the problem complexity. (2) the time distance between phases is not uniformly distributed around the whole period. The number of nodes firing at each phase strongly affects the amount of time between the phases.

\section{FrogSim: An Algorithm for Distributed Graph Coloring}
\label{sec:algorithm}

Although the FrogSim algorithm will be described in terms of an algorithm applied in static sensor networks, it can be applied with very few modifications in any other communication network. The algorithm works iteratively using communication rounds. A communication round corresponds to the calling period ($2\pi$) as known from the models presented in the previous section. The only difference is that the length of a communication round is considered to be one time unit. Therefore, the numerical length of a communication round is denoted by 1, instead of $2\pi$. Each sensor node executes exactly one sensor event in each communication round. The moment in time when a sensor node $i \in V$ executes its sensor event is denoted by $\theta_i \in [0,1)$. Note that $\theta_i$ corresponds to the phase of an oscillator from the models presented in the previous section. Apart from $\theta_i$, a sensor node $i$ also stores its current color, denoted by $c_i \in \mathbb{N}$. For simplicity and without loss of generality, we assume that each color is uniquely identified by a natural number. Thereafter we will use natural numbers greater than zero to refer to colors. Moreover, a sensor event includes the sending of exactly one message. Therefore, each sensor node $i$ maintains a message queue $M_i$ for sensor event messages received from other sensor nodes since the last execution of its own sensor event. The pseudo-code of a sensor event is shown in Algorithm~\ref{algo:coloring-task}. In the following we give a rough description of the algorithm. Detailed technical explanations of the functions of Algorithm~\ref{algo:coloring-task} will be provided later on. 

\begin{algorithm}[!t]
\caption{Sensor event of node $i$}
\label{algo:coloring-task}
\begin{algorithmic}[1]
    \IF{less than $K$ communication rounds executed}
	\STATE $\theta_i := {\sf recalculateTheta()}$
	\STATE $c_i := {\sf minimumColorNotUsed()}$
	\STATE {\sf sendColoringMessage()}
	\STATE $\alpha_i := \alpha_i / \rho$
    \ELSE
	\IF{first communication round of Phase II}
            \STATE {\bf if} ($c_i =1$) {\bf then} $p_i := {\sf randomPositiveInteger()}$
            \STATE {\bf else} $p_i := 0$ {\bf endif}
	%\ELSIF{$\exists m \in M_i \mid (\text{color}_m = c_i \wedge \text{power}_m \geq p_i)$}
	\ELSIF{$\exists m \in M_i \mid (\text{power}_m \geq p_i)$}
		\STATE $c_i :=$ {\sf minimumColorNotUsedByNeighborsWithHigherPower()}
		\STATE $p_i :=$ {\sf adoptPowerFromStrongestNode()}
	\ENDIF
	\STATE {\sf sendRefinementMessage()}
\ENDIF

\STATE {\sf clearMessageQueue()}
\end{algorithmic}
\end{algorithm}

Before the algorithm can be started, it is actually necessary to determine a virtual tree-shaped topology over the sensor network. This task is achieved by using any method from the literature to generate a \emph{minimum spanning tree} in a distributed manner (see, for example,~\cite{gallager1983distributed, awer1987dmst, garay1998sublinear, elkin2006faster}). This tree will determine a single root node that will become a distinguished node of the network (also called the master) with some additional functionalities in comparison to the rest of the nodes. Once this tree has been created, the master node runs a protocol to measure the number of hops (that is, communication rounds) necessary to reach the farthest node in the network. Note that this measure corresponds to the height of the tree. Next, the master node uses this tree to broadcasts an alert to start running the FrogSim algorithm, that is, the first communication round is triggered. This message also includes the height of the tree which will be used later on by each node to define the amount of information that it must store. The simulation of the FrogSim algorithm is composed of two distinct phases. The first phase (called phase I; see lines 1--5 of Algorithm~\ref{algo:coloring-task}) makes use of the model for the desynchronization of frog calling as introduced by Aihara et al.~\cite{aihara2008mathematical}, with only a few modifications. The main difference to other distributed graph coloring algorithms inspired by this model is as follows. The $\theta_i$ values are used for determining the order in which the nodes are allowed to choose colors, whereas in previous algorithms these values were directly associated to specific colors. Note that our algorithm produces a valid coloring already in the first communication round. The second phase (called phase II, see lines 7--15 of Algorithm~\ref{algo:coloring-task}), which is initiated after $K > 0$ communication rounds of phase I, serves to improve the current coloring by means of a refinement technique, similar to distributed local search. \\

Phases I and II of FrogSim will be described in detail in Sections~\ref{sec:phase1} and~\ref{sec:phase2}. Moreover, we will outline how the initially computed tree structure will be used to communicate and store the best coloring found by the algorithm. In this process, each node collects the color identifiers used by its children, determines the highest color used, and sends this information to its parent node. In those cases in which the master node recognizes that the number of colors used in a certain communication round improves over the currently best solution it notifies all the other nodes. This procedure is explained in detail in Section~\ref{sec:data-gathering}.

\subsection{Phase I of FrogSim}
\label{sec:phase1}

During the first $K > 0$ communication rounds (where $K$ is a parameter of the algorithm) each node $i$, when executing its sensor event, executes lines 2--5 of Algorithm~\ref{algo:coloring-task}. First, node $i$ will examine its message queue $M_i$. If $M_i$ contains more than one message from the same sender node, all these messages apart from the last one are deleted. In general, a message $m \in M_i$ sent in this phase has the following format:
\begin{equation}
m = <\text{theta}_m,\text{color}_m,\text{relevance}_m> \quad,
\end{equation}
where $\text{theta}_m \in [0.1)$ contains the $\theta$-value of the emitter, $\text{color}_m$ is the color currently used by the emitter and $\text{relevance} _m$ is a parameter that depends on the number of messages received by the emitter during the last communication round. This parameter controls the weight that is given by node $i$ to the corresponding message $m$. In particular, less weight is given to messages that were emitted by nodes that are influenced by many other nodes. The intuition for this definition of the weights is that the $\theta$-values of nodes that are little influenced by other nodes should converge first. This facilitates the convergence of the $\theta$-values of highly-influenced nodes, which in turn facilitates that the system reaches a stable situation, a term which refers to a situation in which the $\theta$-values do not change anymore. 

Based on the messages in $M_i$, function {\sf recalculateTheta()} recalculates a new value for $\theta_i$:
\begin{equation}
\theta_i := \theta_i + \alpha_i \sum_{m\in M_i} \text{relevance}_m * inc[\theta_m - \theta_i] \quad,
\end{equation}
where $\alpha_i$ is a parameter used to control the convergence of the system, initially set to $0.5$. In general, the lower the value of $\alpha_i$ the smaller the change applied to $\theta_i$. Moreover,  $\text{inc}[\cdot]$ is a function---corresponding to the phase shift function of Equation~\ref{eq:phase-shift}---that is defined as follows:
\begin{equation}
inc[x]=\left\{
\begin{array}{ll}
x - 0.5 &\text{if } x\geq 0\\
x + 0.5 &\text{if } x < 0\\
\end{array} \right.
\end{equation}
Note that this function replaces the sinus function which was originally used in~\cite{aihara2008mathematical} as the phase shift function. This is because we have noticed that this function leads to a better convergence behavior than the sinus function. Next, node $i$ decides for a possibly new color in function {\sf minimumColorNotUsed()}. Formally, the possible color change by node $i$ can be described as:
\begin{equation}
c_i := {\sf min}\{c \in \mathbb{N} \mid \not\exists m\in M_i \text{ with } \text{color}_m = c\}
\end{equation}
In words, node $i$ chooses among the colors that do not appear in any of the received messages $m \in M_i$, the one with the lowest identifier. Finally, node $i$ sends the following message $m$ (see function {\sf sendColoringMessage()}):
\begin{equation}
m = <\text{theta}_m:=\theta_i,\text{color}_m := c_i,\text{relevance}_m := \frac{1}{|M_i|^2}>
\end{equation}
Moreover, node $i$ decreases the value of $\alpha_i$ (see line 5 of Algorithm~\ref{algo:coloring-task}). Hereby, $\rho$ is a parameter of the algorithm that controls the rate of convergence of the $\theta$-values. Note that once the $\theta$-values have converged the current coloring does not change anymore. To conclude a sensor event, node $i$ deletes all messages from its queue $M_i$ (see function {\sf clearMessageQueue()}), that is, $M_i = \emptyset$.

\subsection{Phase II of FrogSim}
\label{sec:phase2}

After $K > 0$ communication rounds, the sensor event of a node $i$ consists of the execution of lines 6--15 of Algorithm~\ref{algo:coloring-task}. As mentioned before, this phase is used for the refinement of the current coloring, similar to a distributed local search. Note that in this phase the $\theta$-values of the nodes are not changed anymore. Within the scope of phase~II, each node $i$ additionally maintains a so-called power parameter $p_i$. This parameter is initialized in the first communication round of phase~II with a positive random integer for the nodes $i$ with $c_i = 1$, and $0$ for the rest of the nodes. The values of these power parameters are used to resolve conflicts that may arise during the color changes executed in phase~II. In particular, in case two neighboring nodes---that is, two nodes that can communicate---have chosen the same color, the one with the higher power value is allowed to keep it. In fact, the usage of such a parameter performs a distributed coloring starting from many nodes at the same time but assuring that it is as good as if the coloring started from a single node. This node will be chosen randomly among those nodes which have the lowest $\theta$-value in each neighborhood. Further down at the end of this section, a graphic example will illustrate the working of phase~II.

\begin{figure}[t!]
\centering
\subfigure[Fictitious situation after phase~I]{
\psset{xunit=.33mm,yunit=.33mm,runit=.33mm}
\begin{pspicture}(100,100)
\rput(20,20){\circlenode[linewidth=0.5pt,fillstyle=solid,fillcolor=white]{1}{$1_2$}}
\put (3,0){$\theta_4=0.1$}
\rput(80,20){\circlenode[linewidth=0.5pt,fillstyle=solid,fillcolor=white]{2}{$3_0$}}
\put (63,0){$\theta_3=0.8$}
\rput(80,80){\circlenode[linewidth=0.5pt,fillstyle=solid,fillcolor=white]{3}{$2_0$}}
\put (63,96){$\theta_2=0.6$}
\rput(20,80){\circlenode[linewidth=0.5pt,fillstyle=solid,fillcolor=white]{4}{$1_5$}}
\put (3,96){$\theta_1=0.3$}

\ncline[linewidth=0.5pt]{-}{1}{2}
\ncline[linewidth=0.5pt]{-}{2}{3}
\ncline[linewidth=0.5pt]{-}{3}{4}
 \end{pspicture}
}
$\;\;\;\;$
\subfigure[Some actions have created a conflict]{
\psset{xunit=.33mm,yunit=.33mm,runit=.33mm}
\begin{pspicture}(100,100)
\rput(20,20){\circlenode[linewidth=0.5pt,fillstyle=solid,fillcolor=white]{1}{$1_2$}}
\rput(80,20){\circlenode[linewidth=0.5pt,fillstyle=solid,fillcolor=white]{2}{$1_5$}}
\rput(80,80){\circlenode[linewidth=0.5pt,fillstyle=solid,fillcolor=white]{3}{$2_5$}}
\rput(20,80){\circlenode[linewidth=0.5pt,fillstyle=solid,fillcolor=white]{4}{$1_5$}}

\ncline[linewidth=0.5pt]{-}{1}{2}
\ncline[linewidth=0.5pt]{-}{2}{3}
\ncline[linewidth=0.5pt]{-}{3}{4}

\end{pspicture}
}
$\;\;\;\;$
\subfigure[Conflicts are resolved]{
\psset{xunit=.33mm,yunit=.33mm,runit=.33mm}
\begin{pspicture}(100,100)
\rput(20,20){\circlenode[linewidth=0.5pt,fillstyle=solid,fillcolor=white]{1}{$2_5$}}
\rput(80,20){\circlenode[linewidth=0.5pt,fillstyle=solid,fillcolor=white]{2}{$1_5$}}
\rput(80,80){\circlenode[linewidth=0.5pt,fillstyle=solid,fillcolor=white]{3}{$2_5$}}
\rput(20,80){\circlenode[linewidth=0.5pt,fillstyle=solid,fillcolor=white]{4}{$1_5$}}

\ncline[linewidth=0.5pt]{-}{1}{2}
\ncline[linewidth=0.5pt]{-}{2}{3}
\ncline[linewidth=0.5pt]{-}{3}{4}

\end{pspicture}
}
\caption{Example of the working of phase~II of \fsimp. Nodes are labeled with their respective color. The nodes' powers are shown as sub-indices of their colors. Graphic (a) shows a fictitious situation after phase~I. Three colors are used in the current feasible coloring. The fictitious $\theta$-values are as indicated besides the nodes. Note that in phase~II they will not change anymore. Initially the nodes with color $1$ receive a random power greater than $0$ (in this case, $2$, respectively $5$), while the remaining nodes receive a power of $0$. First, the node with highest power forces its neighbor to adopt its power (a color change of the neighbor is not necessary). Then, this neighbor, which has color 2, forces its other neighbor to adopt color 1 and power 5 (see graphic (b)). This creates a conflict. However, due to the fact that power 5 is greater than power 2, the last node is forced to change its color from 1 to 2. Note that the final situation uses one color less than the original one.}
\label{fig:phase2}
\end{figure}
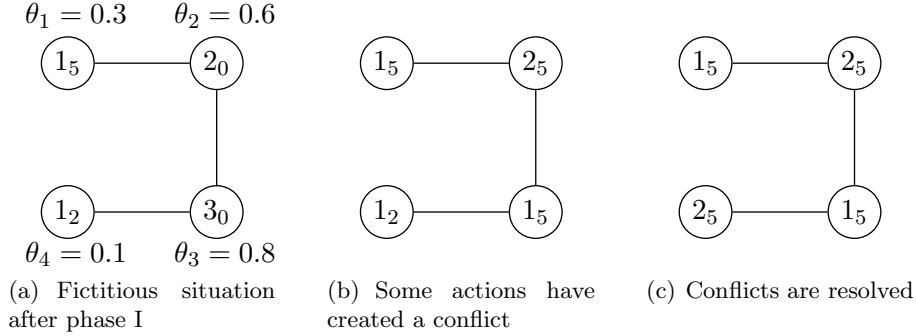

A message $m$ sent by function {\sf sendRefinementMessage()} (see line 14 of Algorithm~\ref{algo:coloring-task}) has the following format:
\begin{equation}
m = <\text{color}_m,\text{power}_m>
\end{equation}
In case the current communication round is not the first communication round of phase~II, node $i$ first examines again its message queue $M_i$. If $M_i$ contains more than one message from the same sender node, all these messages apart from the last one are deleted. Then, the remaining messages are examined, and a color change only occurs if there is a message $m \in M_i$ such that $\text{color}_m = c_i $ and $\text{power}_m \geq p_i$. In words, node $i$ only changes its color if there is an adjacent node with the same color and a higher (or equal) power value. The new color chosen by node $i$ is the first free color that is not already in use by a node influencing node $i$ and that has a power equal to or greater than the power value of node $i$. Formally, the new color $c_i$ is chosen in function {\sf minimumColorNotUsedByNeighborsWithHigherPower()} as follows:
\begin{equation}
c_i := {\sf min}\{\text{c} \in \mathbb{N} \mid \not\exists m\in M_i \text{ with } \text{color}_m = c \wedge \text{power}_m \geq p_i \}
\end{equation}
In addition, node $i$ updates its power value in function {\sf adoptPowerFromStrongestNode()} in the following way:
\begin{equation}
%p_i := {argmax}_{m\in M_i}\{\text{power}_m \mid \text{color}_m < c_i \}
p_i := {argmax}_{m\in M_i}\{\text{power}_m\}
\end{equation}
This is the highest power among the powers of the nodes that have forced node $i$ to choose its current color. As a result, in following communication rounds node $i$ will not be forced to change its color, because with the new power it has priority over all nodes with a lower power. Finally, node $i$ sends a refinement message $m$ in function {\sf sendRefinementMessage()}, where $m$ is defined as follows:
\begin{equation}
m = <\text{color}_m := c_i, \text{power} := p_i>
\end{equation}
The last action of the sensor event consists again in deleting all messages from the message queue $M_i$, that is, $M_i = \emptyset$. Figure~\ref{fig:phase2} shows a small example of the kind of conflicts that phase~II is supposed to resolve. 

\subsection{Determining and Storing the Best Coloring Found}
\label{sec:data-gathering}

It is intuitively clear that the current coloring of our system---that is, the coloring defined by colors $c_i$ for all nodes $i$---does not only improve over time. In some communication rounds, especially during the second phase of the algorithm, the new coloring after the choice of new colors might actually be worse then the coloring of the previous communication round. This behavior is very natural, because the search space of a combinatorial optimization problem is characterized by rather many local minima. If we assume that the current solution corresponds to such a local minimum, the only way to find a better solution is to accept worse solutions for some iterations. In the context of metaheuristic algorithms such an action is known as \emph{escaping from a local minimum}~\cite{BlumRoli03:meta}. 

In order to store the best coloring found by our algorithm over the whole simulation time, the following mechanism is used. Remember that the first action of the algorithm (before simulating phases~I and~II) consisted in the generation of a virtual minimum spanning tree over the network, resulting in a root node (the master). This tree is characterized by its height $h$, which corresponds to the maximum number of communication rounds that a broadcast message sent by the root node needs in order to reach all nodes of the network. In this context, note that $h$ may be minimized by using a priori some methods from the literature which are able to generate spanning trees with minimum diameter in a distributed manner~\cite{bui2004distributed}. 

Each node is required to store its colors from the last $2h$ communication rounds. Moreover, we assume that each node stores the color it has used in the best-found coloring in a specific variable. The way in which this best-found coloring is determined is as follows. First, at each communication round a node sends the maximum color used by itself and its children (with respect to the tree) to its parent in the tree. Such a message only contains two integers (the maximum color and the communication round identifier). Moreover, no additional messages are required because this information can easily be added to the messages that are sent anyway (see lines 4 and 14 of Algorithm~\ref{algo:coloring-task}). Given the height $h$ of the tree, it takes $h$ communication rounds until all the information regarding a specific communication round has reached the root node. Moreover, the number of colors used at this communication round is the maximum color identifier that reaches the root node via one of its children. In case this maximum color is lower than the number of colors used in the currently best-found coloring, the root node broadcasts a message with the corresponding communication round identifier in which this coloring was obtained. In order for this information to reach all the nodes of the network, another $h$ communication rounds are necessary. This is why all nodes must store their colors from last $2h$ communication rounds. Note that these alert messages from the root node can also be propagated using the normal messages of Algorithm~\ref{algo:coloring-task}.

\section{Experimental Results\label{sec:results}}

We coded our algorithm by means of discrete event simulation, implemented from scratch in C++. For the experimental evaluation we chose a large set of different graph topologies: random geometric graphs of different densities, grid graphs of different sizes, and most of the graphs used for the DIMACS challenge~\cite{dimacschallenge}. All graphs that we used for the experimental evaluation can be found for download at \emph{http://www.lsi.upc.edu/\textasciitilde hhernandez/graphcoloring}. Note that an edge connecting two nodes indicates that both nodes are able to communicate directly with each other via their radio antennas.

For the purpose of comparison we re-implemented one of the currently best algorithms from the literature. This algorithm was presented by Finocchi et al. in~\cite{finocchi2005experimental}. For simplicity, this algorithm will henceforth be referred to by {\sf Finocchi}. Unfortunately, the description of this algorithm in the original article contains some ambiguities, which required us to make some decisions regarding certain aspects in the context of the re-implementation. Fortunately, our own implementation of the Finocci algorithm provides generally better results than the ones reported in~\cite{finocchi2005experimental}. This can be verified by comparing the results of the original implementation with the results of our re-implementation for the graph topologies that are used both in~\cite{finocchi2005experimental} and in the present paper. \\

In the following we present the results of three algorithms: (1) Finocci~\cite{finocchi2005experimental}, (2) $\text{FrogSim}^{\varominus}$, which is the FrogSim algorithm without phase~II, and (3) FrogSim, which is the complete FrogSim algorithm. In our opinion, the study of the results of $\text{FrogSim}^{\varominus}$ is worthwhile, because it reflects the power of the frog-based model without any additional improvements of the refinement phase. We applied each of these three stochastic algorithms 100 times to each graph topology and report the best coloring found in all 100 runs, as well as the average quality of the best colorings found per run. The number of rounds necessary to reach these solutions is---due to space reasons---not included in the result tables. However, it is important to note that algorithms such as Finocci and FrogSim, when used in sensor networks, are generally carried out continuously in a lower-level layer of the network. Therefore, the number of communication rounds necessary to reach the best solution are not that significant. Instead our algorithm continually tries to improve the current solution. As an informative note, our algorithm requires, on average, $10.34$ communication rounds for finding its best solution in phase~I. After entering phase~II the best solution is reached, on average, after $3.46$ communication rounds. In total, FrogSim requires, on average, $24.33$ communication rounds for finding its best solution. The algorithm of Finocchi et al.~uses, on average, a comparable number of communication rounds ($19.83$). It should be noted that, in the case of FrogSim, these numbers do not depend so much on the size of the network. However, FrogSim takes generally more communication rounds for those graphs that have a larger number of edges. 

After tuning by hand, we decided to use a communication round limit of 100 rounds for FrogSim. Moreover, parameter $K$, which specifies the number of communication rounds for phase~I, was always set to 80. As a last remark, note that the size of the messages used in FrogSim is constant ($O(1)$). In other words, the message size does not depend on the network size. This is surely a desirably property of a distributed algorithm for graph coloring.

\subsection{Results for Random Geometric Graphs}

Random geometric graphs are popular models for sensor networks. Therefore, they are frequently used for the evaluation of algorithms developed for such networks. They are generated by randomly distributing a set of $n$ nodes in the $[0,1]^2$ area. Two vertices $u$ and $v$ are connected by an edge, if and only if $d(u,v)\leq r$, where $d(.,.)$ is the Euclidean distance and $r > 0$ is a threshold. More specifically, the three algorithms were applied to 40 random geometric graphs with $n \in  \{20,50,100,200\}$ and $r = 0.05$. 

Table~\ref{tab:res-random} presents the results obtained for this set of instances. In particular, the first column shows the names of the instances and the second column provides a triple $(n,\Delta,\chi)$, where $n$ is the number of nodes, $\Delta$ the maximum degree, and $\chi$ the chromatic number of the corresponding graph. In case of a question mark, the chromatic number is not known. The following three groups of columns provide the results obtained by the three algorithms. For each algorithm we first give the number of colors from the best coloring found over $100$ independent runs. In the second column, we show the average number of colors used by the $100$ colorings obtained in $100$ runs. For ease of comparison the best performing algorithm for each instance is indicated in bold face. Hereby, the \emph{best performing algorithm} is defined as the algorithm that finds the best coloring. Ties are broken (if possible) by the average values. The four bottom rows of the table provide a summary of the results. The first one of these rows gives averages for each column. In addition, the last three rows summarize how each algorithm is performing in comparison to the others. The first of these rows (labelled {\bf \# times better}) indicates for each algorithm the number of instances for which the corresponding algorithm was the sole winner, that is, better than the other two algorithms. The second row (labelled {\bf \# times all equal}) indicates for how many instances the results of the three algorithms were equal, whereas the last table row indicates for each algorithm the number of instances for which the corresponding algorithm was the sole looser.

\begin{table*}[!t]
\caption{Results for random geometric graphs.}\label{tab:res-random}
\centering
\scalebox{0.7}{
\begin{tabular}{ccccrrcrrcrr}
\hline \hline
\multirow{2}{*}{Instance} & $\;\;$ & \multirow{2}{*}{$(n,\Delta,\chi)$} & $\;\;$ & \multicolumn{2}{c}{Finocchi}& $\;\;$ & \multicolumn{2}{c}{$\text{FrogSim}^{\varominus}$} & $\;\;$ &\multicolumn{2}{c}{$\text{FrogSim}$}\\
&& && colors & avg. && colors & avg. && colors & avg. \\
\cline{1-1} \cline{3-3} \cline{5-6} \cline{8-9} \cline{11-12}
random-graph-n20-r05-1.gph&&(20,2,?) && {\bf 2}  & 2.000 && {\bf 2}  & 2.000 && {\bf 2}  & 2.000\\
random-graph-n20-r05-10.gph&&(20,1,?) && {\bf 2}  & 2.000 && {\bf 2}  & 2.000 && {\bf 2}  & 2.000\\
random-graph-n20-r05-2.gph&&(20,2,?) && {\bf 3}  & 3.000 && {\bf 3}  & 3.000 && {\bf 3}  & 3.000\\
random-graph-n20-r05-3.gph&&(20,2,?) && {\bf 2}  & 2.000 && {\bf 2}  & 2.000 && {\bf 2}  & 2.000\\
random-graph-n20-r05-4.gph&&(20,3,?) && {\bf 3}  & 3.000 && {\bf 3}  & 3.000 && {\bf 3}  & 3.000\\
random-graph-n20-r05-5.gph&&(20,3,?) && {\bf 3}  & 3.000 && {\bf 3}  & 3.000 && {\bf 3}  & 3.000\\
random-graph-n20-r05-6.gph&&(20,1,?) && {\bf 2}  & 2.000 && {\bf 2}  & 2.000 && {\bf 2}  & 2.000\\
random-graph-n20-r05-7.gph&&(20,2,?) && {\bf 3}  & 3.000 && {\bf 3}  & 3.000 && {\bf 3}  & 3.000\\
random-graph-n20-r05-8.gph&&(20,2,?) && {\bf 3}  & 3.000 && {\bf 3}  & 3.000 && {\bf 3}  & 3.000\\
random-graph-n20-r05-9.gph&&(20,2,?) && {\bf 2}  & 2.000 && {\bf 2}  & 2.000 && {\bf 2}  & 2.000\\
random-graph-n50-r05-1.gph&&(50,6,?) && {\bf 6}  & 6.000 && {\bf 6}  & 6.000 && {\bf 6}  & 6.000\\
random-graph-n50-r05-10.gph&&(50,6,?) && {\bf 5}  & 5.000 && {\bf 5}  & 5.000 && {\bf 5}  & 5.000\\
random-graph-n50-r05-2.gph&&(50,3,?) && {\bf 3}  & 3.000 && {\bf 3}  & 3.000 && {\bf 3}  & 3.000\\
random-graph-n50-r05-3.gph&&(50,4,?) && {\bf 4}  & 4.000 && {\bf 4}  & 4.000 && {\bf 4}  & 4.000\\
random-graph-n50-r05-4.gph&&(50,4,?) && 3 & 3.260 && {\bf 3}  & 3.000 && {\bf 3}  & 3.000\\
random-graph-n50-r05-5.gph&&(50,4,?) && {\bf 3}  & 3.000 && {\bf 3}  & 3.000 && {\bf 3}  & 3.000\\
random-graph-n50-r05-6.gph&&(50,4,?) && {\bf 4}  & 4.000 && {\bf 4}  & 4.000 && {\bf 4}  & 4.000\\
random-graph-n50-r05-7.gph&&(50,6,?) && {\bf 4}  & 4.000 && {\bf 4}  & 4.000 && {\bf 4}  & 4.000\\
random-graph-n50-r05-8.gph&&(50,4,?) && {\bf 3}  & 3.000 && {\bf 3}  & 3.000 && {\bf 3}  & 3.000\\
random-graph-n50-r05-9.gph&&(50,3,?) && {\bf 3}  & 3.000 && {\bf 3}  & 3.000 && {\bf 3}  & 3.000\\
random-graph-n100-r05-1.gph&&(100,8,?) && {\bf 5}  & 5.000 && {\bf 5}  & 5.000 && {\bf 5}  & 5.000\\
random-graph-n100-r05-10.gph&&(100,8,?) && {\bf 5}  & 5.000 && 5 & 5.820 && 5 & 5.220\\
random-graph-n100-r05-2.gph&&(100,7,?) && 4 & 4.420 && 4 & 4.430 && {\bf 4}  & 4.000\\
random-graph-n100-r05-3.gph&&(100,7,?) && {\bf 6}  & 6.000 && {\bf 6}  & 6.000 && {\bf 6}  & 6.000\\
random-graph-n100-r05-4.gph&&(100,9,?) && {\bf 5}  & 5.000 && 5 & 5.560 && 5 & 5.410\\
random-graph-n100-r05-5.gph&&(100,7,?) && 4 & 4.500 && 4 & 4.470 && {\bf 4}  & 4.000\\
random-graph-n100-r05-6.gph&&(100,6,?) && {\bf 6}  & 6.000 && {\bf 6}  & 6.000 && {\bf 6}  & 6.000\\
random-graph-n100-r05-7.gph&&(100,6,?) && 5 & 5.000 && 4 & 4.450 && {\bf 4}  & 4.200\\
random-graph-n100-r05-8.gph&&(100,6,?) && 5 & 5.000 && 4 & 4.110 && {\bf 4}  & 4.000\\
random-graph-n100-r05-9.gph&&(100,7,?) && {\bf 6}  & 6.000 && {\bf 6}  & 6.000 && {\bf 6}  & 6.000\\
random-graph-n200-r05-1.gph&&(200,13,?) && 10 & 10.000 && 8 & 8.500 && {\bf 8}  & 8.360\\
random-graph-n200-r05-10.gph&&(200,13,?) && {\bf 8}  & 8.000 && {\bf 8}  & 8.000 && {\bf 8}  & 8.000\\
random-graph-n200-r05-2.gph&&(200,12,?) && {\bf 8}  & 8.000 && 8 & 8.030 && {\bf 8}  & 8.000\\
random-graph-n200-r05-3.gph&&(200,12,?) && 8 & 8.000 && 7 & 7.640 && {\bf 7}  & 7.490\\
random-graph-n200-r05-4.gph&&(200,12,?) && 9 & 9.000 && 8 & 8.100 && {\bf 8}  & 8.000\\
random-graph-n200-r05-5.gph&&(200,17,?) && 10 & 10.000 && 8 & 8.990 && {\bf 8}  & 8.840\\
random-graph-n200-r05-6.gph&&(200,12,?) && 8 & 8.260 && {\bf 8}  & 8.000 && {\bf 8}  & 8.000\\
random-graph-n200-r05-7.gph&&(200,12,?) && 7 & 7.000 && 6 & 6.830 && {\bf 6}  & 6.750\\
random-graph-n200-r05-8.gph&&(200,11,?) && 8 & 8.660 && 7 & 7.630 && {\bf 7}  & 7.490\\
random-graph-n200-r05-9.gph&&(200,11,?) && {\bf 7}  & 7.000 && 7 & 7.260 && 7 & 7.050\\
\cline{1-1} \cline{3-3} \cline{5-6} \cline{8-9} \cline{11-12}
average && && 4.925 & 4.978 && 4.675 & 4.845 && 4.675 & 4.770\\
\cline{1-1} \cline{3-3} \cline{5-6} \cline{8-9} \cline{11-12}
\# times better && && \multicolumn{2}{c}{3} && \multicolumn{2}{c}{0} && \multicolumn{2}{c}{10}\\
\# times all equal && &&\multicolumn{2}{c}{24} && \multicolumn{2}{c}{24} && \multicolumn{2}{c}{24}\\
\# times worse && &&\multicolumn{2}{c}{11} && \multicolumn{2}{c}{5} && \multicolumn{2}{c}{0}\\
\hline \hline
\end{tabular}}
\end{table*}

As expected, the results show that the smaller the size of the graph, the easier it is to find good colorings. The algorithms obtain equivalent results for $24$ out of $40$ instances (note that all small instances with $20$ and $50$ nodes are included in this set). Although \fin\ is $3$ times better than the other two algorithms it is also worse in $11$ topologies. More importantly, \fin\ is not always able to match the FrogSim algorithms in terms of the best colorings for each instance. More specifically, \fin\ uses $0.250$ colors more on average than both FrogSim algorithms. Although \fsim\ is not able to outperform the other two algorithms for any given instance it only obtains the worst result for $5$ instances. \fsimp\ improves over the results of \fsim\ especially for the larger instances. It turns out to be the sole winner for $10$ instances. It is interesting to note that in those cases where \fsimp\ is better than \fsim\ this is due to the average solution quality. In this sense it can be said that in the context of random geometric graphs the use of phase~II makes the FrogSim algorithm more robust. It is also important to note that the best colorings obtained are---for almost all instances---better than $\Delta+1$ colors.

\begin{figure}[!t]
\centering
 \includegraphics[width=5cm,angle=-90]{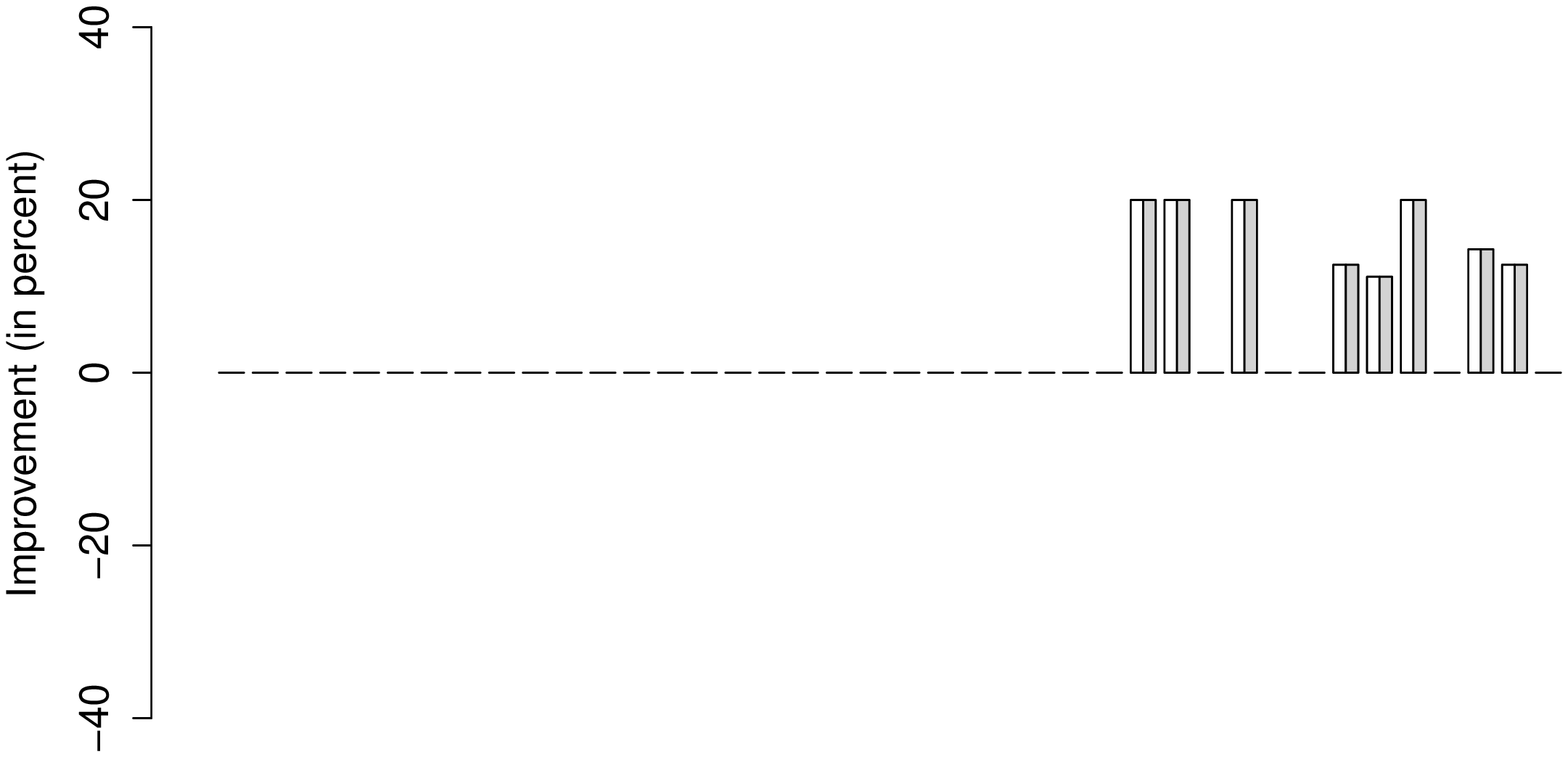}
 \includegraphics[width=5cm,angle=-90]{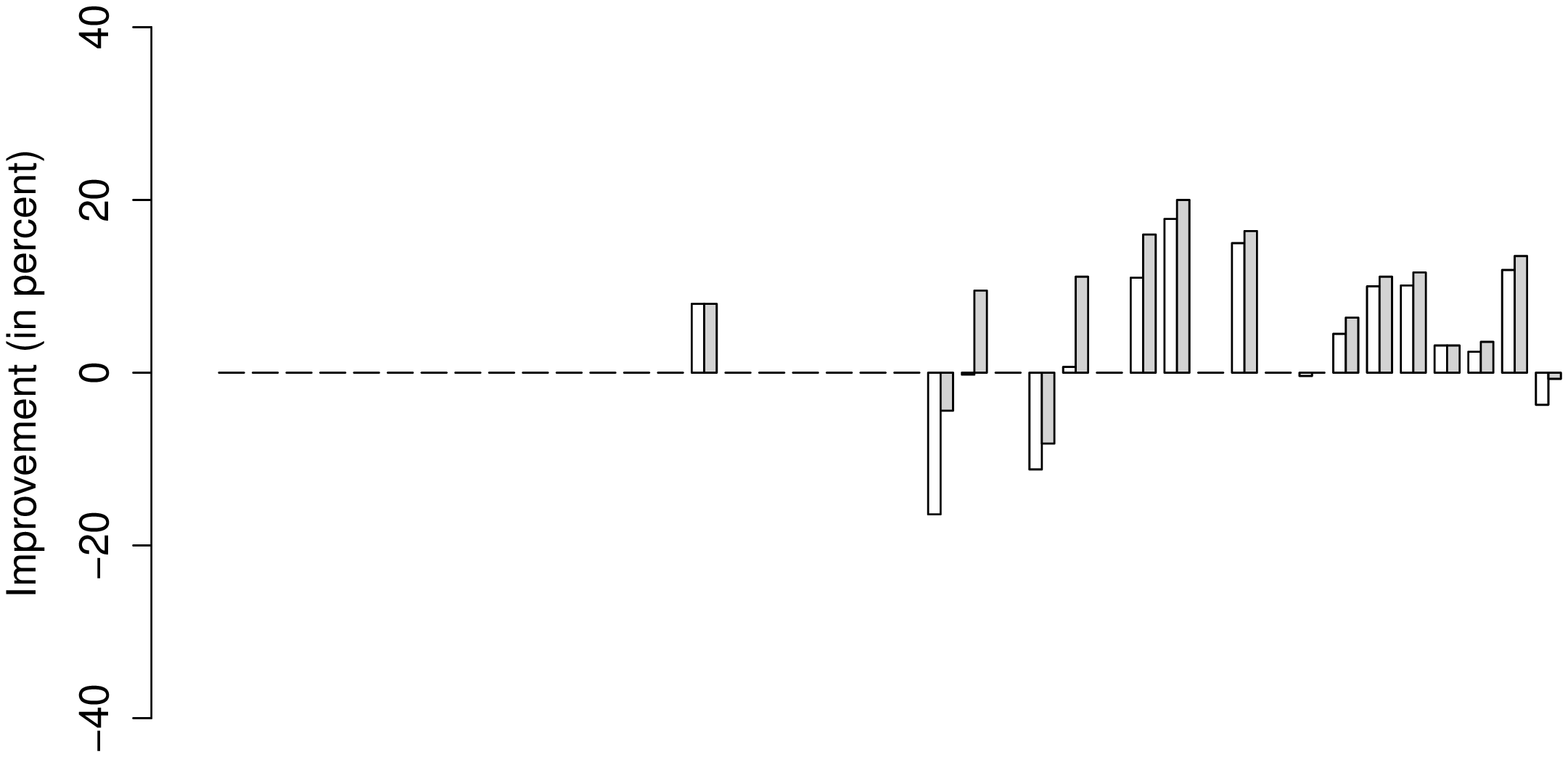}
\caption{Summary of results for random geometric graphs. Both graphics show the performance improvement of \fsim\ (light gray bars) and \fsimp\ (dark gray bars) over \fin\ (in percent). The instances of Table~\ref{tab:res-random} are treated from left to right in the same order. The top graphic concerns the best colorings found, whereas the bottom graphic concerns the average solution quality.}
\label{fig:res-random}
\end{figure}

In addition to Table~\ref{tab:res-random}, the results are also presented in a visual form in Figure~\ref{fig:res-random}. For each graph (x-axis) the  improvement of \fsim\ and \fsimp\ over \fin\ in terms of the best coloring (top graphic) and the average solution quality (bottom graphic) is presented. The 40 considered graphs are ordered from left to right as they appear in Table~\ref{tab:res-random}. These graphics show nicely that the FrogSim algorithms gain an advantage over \fin\ with growing instance size (from left to right). The bottom graphic shows that there are only three graphs for which \fin\ achieves a better average solution quality. 

\subsection{Results for DIMACS Graphs}
\label{sec:results-dimacs}

One of the most popular sets of instances in the context of graph coloring is the one introduced for the \emph{second DIMACS challenge}~\cite{dimacschallenge}. This challenge had among its objectives to establish the state-of-the-art techniques for centralized graph coloring. These graphs are generally larger and more complex than, for example, random geometric graphs. The instances originate from very different contexts, ranging from industrial problems to hand-crafted cases that were created to show the ineffectiveness of certain algorithms. This set of instances is often used as a benchmark to study the quality of new algorithms, also in the context of distributed graph coloring (see, for example,~\cite{finocchi2005experimental,lee2008firefly,malaguti2010survey,lu2010memetic}). \\

The results are presented in Tables~\ref{tab:res-dimacs1} and~\ref{tab:res-dimacs2}, in the same way as in the case of random geometric graphs. Concerning the chromatic numbers, in many cases they are known. In the cases in which they are not known, we either provide an upper bound (in the form $\leq$X) or a question mark. As a general remark before analyzing the results in depth, we would like to mention that for distributed algorithms it is very difficult, if not impossible, to capture the global structure of these graphs in many cases. Therefore, it is not surprising that the results obtained by distributed algorithms are often far away from the chromatic numbers.

\begin{table*}[!t]
\caption{Results for the first set of instances from the DIMACS challenge.}\label{tab:res-dimacs1}
\centering
\scalebox{0.7}{
\begin{tabular}{ccccrrcrrcrr}
\hline \hline
\multirow{2}{*}{Instance} & $\;\;$ & \multirow{2}{*}{$(n,\Delta,\chi)$} & $\;\;$ & \multicolumn{2}{c}{Finocchi}& $\;\;$ & \multicolumn{2}{c}{$\text{FrogSim}^{\varominus}$} & $\;\;$ &\multicolumn{2}{c}{$\text{FrogSim}$}\\
&& && colors & avg. && colors & avg. && colors & avg. \\
\cline{1-1} \cline{3-3} \cline{5-6} \cline{8-9} \cline{11-12}
DSJC1000.1.col&&(1000,127,$\leq$20) && 30 & 31.250 && {\bf 29}  & 29.564 && {\bf 29}  & 29.564\\
DSJC1000.5.col&&(1000,551,$\leq$83) && 124 & 126.550 && {\bf 118}  & 120.505 && {\bf 118}  & 120.505\\
DSJC1000.9.col&&(1000,924,$\leq$224) && 315 & 321.140 && {\bf 297}  & 303.594 && {\bf 297}  & 303.594\\
DSJC125.1.col&&(125,23,?) && 8 & 8.000 && 7 & 7.485 && {\bf 7}  & 7.386\\
DSJC125.5.col&&(125,75,?) && 24 & 25.630 && 22 & 23.535 && {\bf 22}  & 23.475\\
DSJC125.9.col&&(125,120,?) && 54 & 55.300 && 50 & 53.030 && {\bf 50}  & 53.020\\
DSJC250.1.col&&(250,38,?) && 12 & 12.750 && 11 & 11.941 && {\bf 11}  & 11.931\\
DSJC250.5.col&&(250,147,?) && 40 & 42.420 && 38 & 39.792 && {\bf 38}  & 39.772\\
DSJC250.9.col&&(250,234,?) && 95 & 97.290 && {\bf 89}  & 92.297 && {\bf 89}  & 92.297\\
DSJC500.1.col&&(500,68,$\leq$12) && 18 & 19.310 && 17 & 18.218 && {\bf 17}  & 18.178\\
DSJC500.5.col&&(500,286,$\leq$48) && 70 & 72.800 && {\bf 67}  & 68.762 && {\bf 67}  & 68.762\\
DSJC500.9.col&&(500,471,$\leq$126) && 170 & 177.000 && {\bf 164}  & 167.703 && {\bf 164}  & 167.703\\
DSJR500.1.col&&(500,25,?) && 14 & 14.540 && 13 & 13.960 && {\bf 13}  & 13.901\\
DSJR500.1c.col&&(500,497,$\leq$85) && 100 & 108.290 && 97 & 103.129 && {\bf 97}  & 102.980\\
DSJR500.5.col&&(500,388,$\leq$122) && 142 & 146.770 && 141 & 146.337 && {\bf 140}  & 144.634\\
flat1000-50-0.col&&(1000,520,50) && 121 & 124.420 && {\bf 116}  & 118.139 && {\bf 116}  & 118.139\\
flat1000-60-0.col&&(1000,524,60) && 121 & 124.730 && {\bf 115}  & 118.604 && {\bf 115}  & 118.604\\
flat1000-76-0.col&&(1000,532,76) && 121 & 125.220 && {\bf 117}  & 119.119 && {\bf 117}  & 119.119\\
flat300-20-0.col&&(300,160,20) && 44 & 46.400 && 42 & 43.485 && {\bf 42}  & 43.455\\
flat300-26-0.col&&(300,158,26) && 46 & 47.660 && 42 & 44.198 && {\bf 42}  & 44.188\\
flat300-28-0.col&&(300,162,28) && 45 & 47.260 && {\bf 43}  & 44.366 && {\bf 43}  & 44.366\\
fpsol2.i.1.col&&(496,252,65) && {\bf 65}  & 65.000 && {\bf 65}  & 65.000 && {\bf 65}  & 65.000\\
fpsol2.i.2.col&&(451,346,30) && 30 & 30.360 && 30 & 30.178 && {\bf 30}  & 30.030\\
fpsol2.i.3.col&&(425,346,30) && 30 & 30.450 && 30 & 30.109 && {\bf 30}  & 30.059\\
inithx.i.1.col&&(864,502,54) && {\bf 54}  & 54.000 && {\bf 54}  & 54.000 && {\bf 54}  & 54.000\\
inithx.i.2.col&&(645,541,31) && 31 & 31.020 && {\bf 31}  & 31.000 && {\bf 31}  & 31.000\\
inithx.i.3.col&&(621,542,31) && {\bf 31}  & 31.000 && {\bf 31}  & 31.000 && {\bf 31}  & 31.000\\
le450-15a.col&&(450,99,15) && 21 & 21.930 && 20 & 21.010 && {\bf 20}  & 20.733\\
le450-15b.col&&(450,94,15) && 20 & 21.440 && 20 & 21.059 && {\bf 20}  & 20.693\\
le450-15c.col&&(450,139,15) && 29 & 30.580 && 28 & 29.535 && {\bf 28}  & 29.257\\
le450-15d.col&&(450,138,15) && 29 & 30.510 && 28 & 29.545 && {\bf 28}  & 29.366\\
le450-25a.col&&(450,128,25) && 27 & 28.830 && 27 & 27.832 && {\bf 26}  & 27.416\\
le450-25b.col&&(450,111,25) && 26 & 27.660 && 26 & 27.317 && {\bf 26}  & 26.941\\
le450-25c.col&&(450,179,25) && 35 & 35.890 && 34 & 35.317 && {\bf 33}  & 34.861\\
le450-25d.col&&(450,157,25) && 35 & 35.650 && 33 & 35.406 && {\bf 33}  & 34.851\\
le450-5a.col&&(450,42,5) && 13 & 13.230 && 12 & 12.129 && {\bf 11}  & 12.069\\
le450-5b.col&&(450,42,5) && 12 & 13.220 && 12 & 12.030 && {\bf 12}  & 12.020\\
le450-5c.col&&(450,66,5) && 15 & 16.590 && 11 & 13.218 && {\bf 11}  & 13.178\\
le450-5d.col&&(450,68,5) && 15 & 16.580 && 11 & 13.347 && {\bf 11}  & 13.327\\
\cline{1-1} \cline{3-3} \cline{5-6} \cline{8-9} \cline{11-12}
average && && 57.231 & 59.197 && 54.821 & 56.584 && 54.718 & 56.446\\
\cline{1-1} \cline{3-3} \cline{5-6} \cline{8-9} \cline{11-12}
\# times better && && \multicolumn{2}{c}{0} && \multicolumn{2}{c}{0} && \multicolumn{2}{c}{25}\\
\# times all equal && &&\multicolumn{2}{c}{3} && \multicolumn{2}{c}{3} && \multicolumn{2}{c}{3}\\
\# times worse && &&\multicolumn{2}{c}{36} && \multicolumn{2}{c}{0} && \multicolumn{2}{c}{0}\\
\hline \hline
\end{tabular}}
\end{table*}

\begin{table*}[!t]
\caption{Results for the second set of instances from the DIMACS challenge.}\label{tab:res-dimacs2}
\centering
\scalebox{0.7}{
\begin{tabular}{ccccrrcrrcrr}
\hline \hline
\multirow{2}{*}{Instance} & $\;\;$ & \multirow{2}{*}{$(n,\Delta,\chi)$} & $\;\;$ & \multicolumn{2}{c}{Finocchi}& $\;\;$ & \multicolumn{2}{c}{$\text{FrogSim}^{\varominus}$} & $\;\;$ &\multicolumn{2}{c}{$\text{FrogSim}$}\\
&& && colors & avg. && colors & avg. && colors & avg. \\
\cline{1-1} \cline{3-3} \cline{5-6} \cline{8-9} \cline{11-12}
anna.col&&(138,71,11) && {\bf 11}  & 11.000 && {\bf 11}  & 11.000 && {\bf 11}  & 11.000\\
david.col&&(87,82,11) && 11 & 11.720 && 11 & 11.446 && {\bf 11}  & 11.297\\
games120.col&&(120,13,9) && {\bf 9}  & 9.000 && 9 & 9.040 && {\bf 9}  & 9.000\\
homer.col&&(561,99,13) && 14 & 14.070 && 13 & 13.644 && {\bf 13}  & 13.158\\
huck.col&&(74,53,11) && {\bf 11}  & 11.000 && {\bf 11}  & 11.000 && {\bf 11}  & 11.000\\
jean.col&&(80,36,10) && {\bf 10}  & 10.000 && 10 & 10.069 && {\bf 10}  & 10.000\\
miles1000.col&&(128,86,42) && 43 & 44.990 && 43 & 44.327 && {\bf 42}  & 44.000\\
miles1500.col&&(128,106,73) && 74 & 74.220 && 73 & 73.861 && {\bf 73}  & 73.614\\
miles250.col&&(128,16,8) && 9 & 10.160 && 8 & 8.782 && {\bf 8}  & 8.683\\
miles500.col&&(128,38,20) && 21 & 22.120 && 20 & 21.297 && {\bf 20}  & 21.139\\
miles750.col&&(128,64,31) && 32 & 33.330 && 31 & 33.050 && {\bf 31}  & 32.832\\
mulsol.i.1.col&&(197,121,49) && {\bf 49}  & 49.000 && {\bf 49}  & 49.000 && {\bf 49}  & 49.000\\
mulsol.i.2.col&&(188,156,31) && 31 & 31.360 && {\bf 31}  & 31.000 && {\bf 31}  & 31.000\\
mulsol.i.3.col&&(184,157,31) && 31 & 31.140 && {\bf 31}  & 31.000 && {\bf 31}  & 31.000\\
mulsol.i.4.col&&(185,158,31) && 31 & 31.060 && {\bf 31}  & 31.000 && {\bf 31}  & 31.000\\
mulsol.i.5.col&&(186,159,31) && 31 & 31.330 && {\bf 31}  & 31.000 && {\bf 31}  & 31.000\\
myciel2.col&&(5,2,3) && {\bf 3}  & 3.000 && {\bf 3}  & 3.000 && {\bf 3}  & 3.000\\
myciel3.col&&(11,5,4) && 4 & 4.060 && {\bf 4}  & 4.000 && {\bf 4}  & 4.000\\
myciel4.col&&(23,11,5) && 5 & 5.180 && {\bf 5}  & 5.000 && {\bf 5}  & 5.000\\
myciel5.col&&(47,23,6) && 6 & 6.230 && {\bf 6}  & 6.000 && {\bf 6}  & 6.000\\
myciel6.col&&(95,47,7) && 7 & 7.080 && {\bf 7}  & 7.000 && {\bf 7}  & 7.000\\
myciel7.col&&(191,95,8) && 8 & 8.290 && 8 & 8.059 && {\bf 8}  & 8.000\\
queen10-10.col&&(100,35,?) && 15 & 15.420 && 14 & 14.228 && {\bf 14}  & 14.188\\
queen11-11.col&&(121,40,11) && 17 & 17.230 && {\bf 14}  & 15.653 && {\bf 14}  & 15.653\\
queen12-12.col&&(144,43,?) && 17 & 17.700 && 16 & 16.960 && {\bf 16}  & 16.921\\
queen13-13.col&&(169,48,13) && 19 & 19.950 && 17 & 18.188 && {\bf 17}  & 18.178\\
queen14-14.col&&(196,51,?) && 20 & 20.730 && 18 & 19.545 && {\bf 18}  & 19.535\\
queen15-15.col&&(225,56,?) && 21 & 22.160 && {\bf 20}  & 20.762 && {\bf 20}  & 20.762\\
queen16-16.col&&(256,59,?) && 21 & 23.100 && {\bf 21}  & 21.990 && {\bf 21}  & 21.990\\
queen5-5.col&&(25,16,5) && 5 & 6.790 && 7 & 7.238 && {\bf 5}  & 6.752\\
queen6-6.col&&(36,19,7) && 9 & 9.760 && {\bf 8}  & 8.743 && {\bf 8}  & 8.743\\
queen7-7.col&&(49,24,7) && 10 & 10.920 && 10 & 10.079 && {\bf 10}  & 10.000\\
queen8-12.col&&(96,32,12) && 15 & 15.280 && 13 & 14.386 && {\bf 13}  & 14.327\\
queen8-8.col&&(64,27,9) && 11 & 12.330 && {\bf 11}  & 11.752 && {\bf 11}  & 11.752\\
queen9-9.col&&(81,32,10) && 12 & 13.510 && 12 & 13.000 && {\bf 12}  & 12.911\\
school1.col&&(385,282,?) && 40 & 41.800 && 35 & 38.772 && {\bf 35}  & 38.703\\
school1-nsh.col&&(352,232,?) && 37 & 38.780 && 31 & 35.762 && {\bf 31}  & 35.614\\
zeroin.i.1.col&&(211,111,49) && 49 & 49.170 && {\bf 49}  & 49.000 && {\bf 49}  & 49.000\\
zeroin.i.2.col&&(211,140,30) && {\bf 30}  & 30.000 && 30 & 30.010 && 30 & 30.010\\
zeroin.i.3.col&&(206,140,30) && 30 & 30.310 && 30 & 30.010 && {\bf 30}  & 30.000\\
\cline{1-1} \cline{3-3} \cline{5-6} \cline{8-9} \cline{11-12}
average && && 20.725 & 21.357 && 20.050 & 20.741 && 19.975 & 20.669\\
\cline{1-1} \cline{3-3} \cline{5-6} \cline{8-9} \cline{11-12}
\# times better && && \multicolumn{2}{c}{1} && \multicolumn{2}{c}{0} && \multicolumn{2}{c}{19}\\
\# times all equal && &&\multicolumn{2}{c}{4} && \multicolumn{2}{c}{4} && \multicolumn{2}{c}{4}\\
\# times worse && &&\multicolumn{2}{c}{32} && \multicolumn{2}{c}{3} && \multicolumn{2}{c}{0}\\
\hline \hline
\end{tabular}}
\end{table*}

First it should be emphasized that the FrogSim algorithms achieve the best results for all instances except for instance zeroin.i.2.col (see Table~\ref{tab:res-dimacs2}), where \fin\ achieves a slightly better average solution quality. Moreover, only in seven further cases, \fin\ is able to match the results of the FrogSim algorithms. On the other side, for some instances the FrogSim algorithms improve remarkably over \fin. Consider, for example, instance DSJC1000.9.col (see Table~\ref{tab:res-dimacs2}) where the best colorings found by the FrogSim algorithms need 297 colors, while the best coloring found by \fin\ uses 315 colors. Other examples of remarkable improvements over \fin\ are the six flat$*$ instances from Table~\ref{tab:res-dimacs2}. Concerning the comparison between \fsim\ and \fsimp, we can state that the power of the algorithm can clearly be attributed to the first (frog-inspired) phase. As in the case of random geometric graphs, phase~II of FrogSim basically helps to make the algorithm more robust. It should also be emphasized that, in all cases, the FrogSim colorings require a number of colors that is smaller than $\Delta + 1$. Although in most cases the best solution obtained is not an optimal coloring---respectively, we do not know whether it is or not---for most of the instances of type mulsol.X, myciel.X and zeroin.X our algorithm generates optimal colorings in each of the $100$ applications per instance. 

\begin{figure}[!t]
\centering
 \includegraphics[width=5cm,angle=-90]{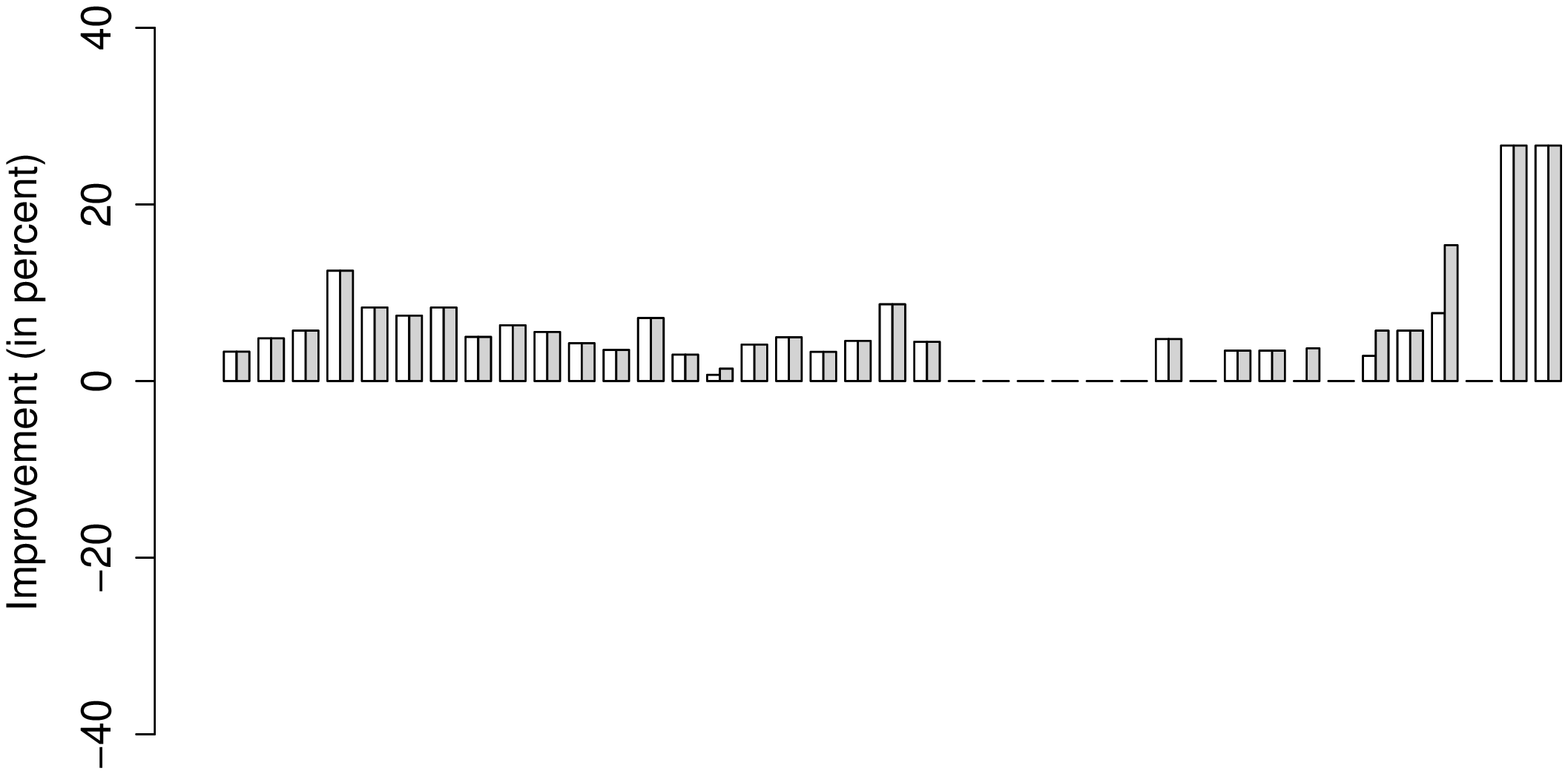}
 \includegraphics[width=5cm,angle=-90]{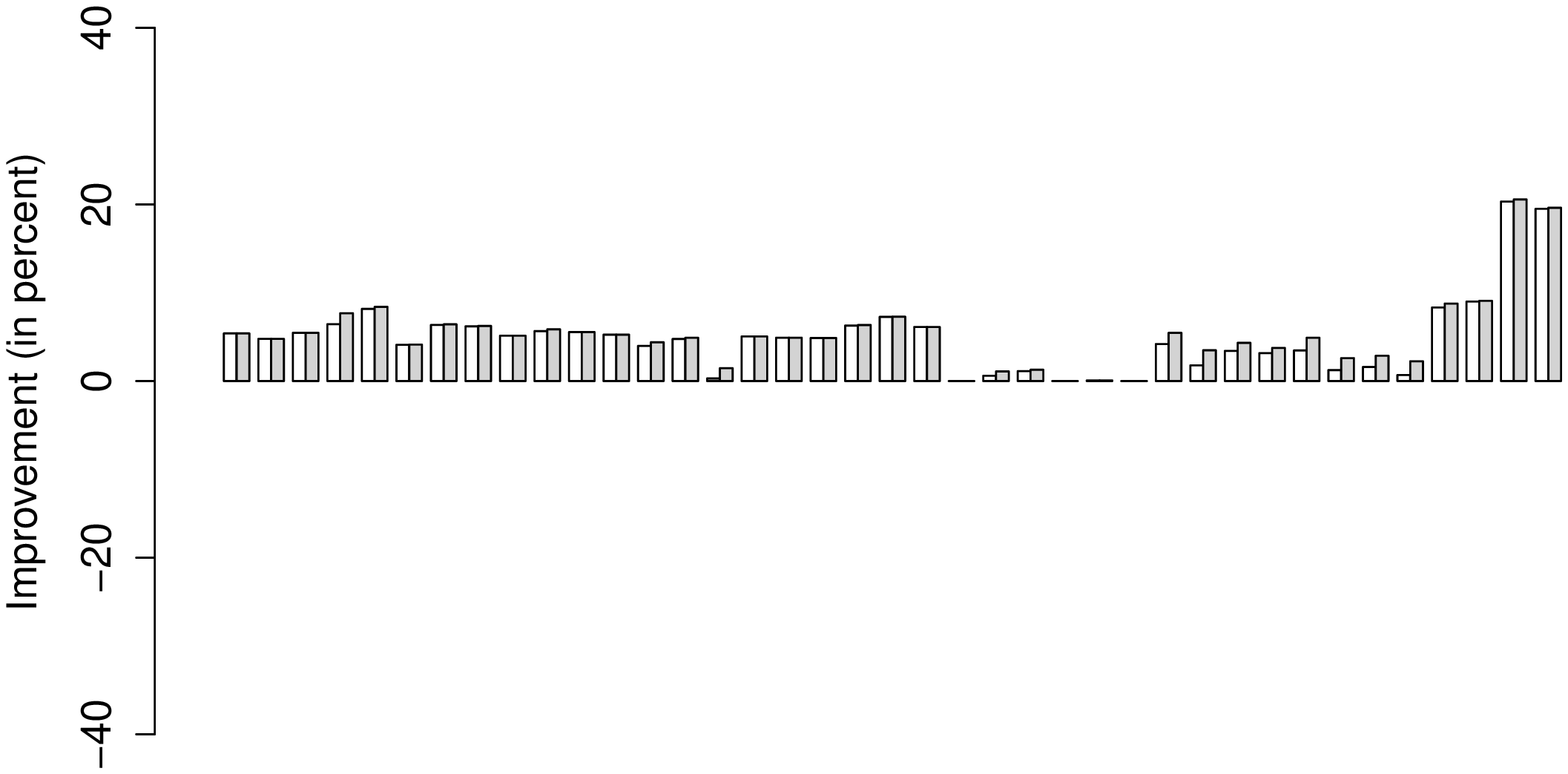}
\caption{Summary of results for the first set of instances from the DIMACS challenge. Both graphics show the performance improvement of \fsim\ (light gray bars) and \fsimp\ (dark gray bars) over \fin\ (in percent). The instances of Table~\ref{tab:res-dimacs1} are treated from left to right in the same order. The top graphic concerns the best colorings found, whereas the bottom graphic concerns the average solution quality.}
\label{fig:res-dimacs1}
\end{figure}

\begin{figure}[!t]
\centering
 \includegraphics[width=5cm,angle=-90]{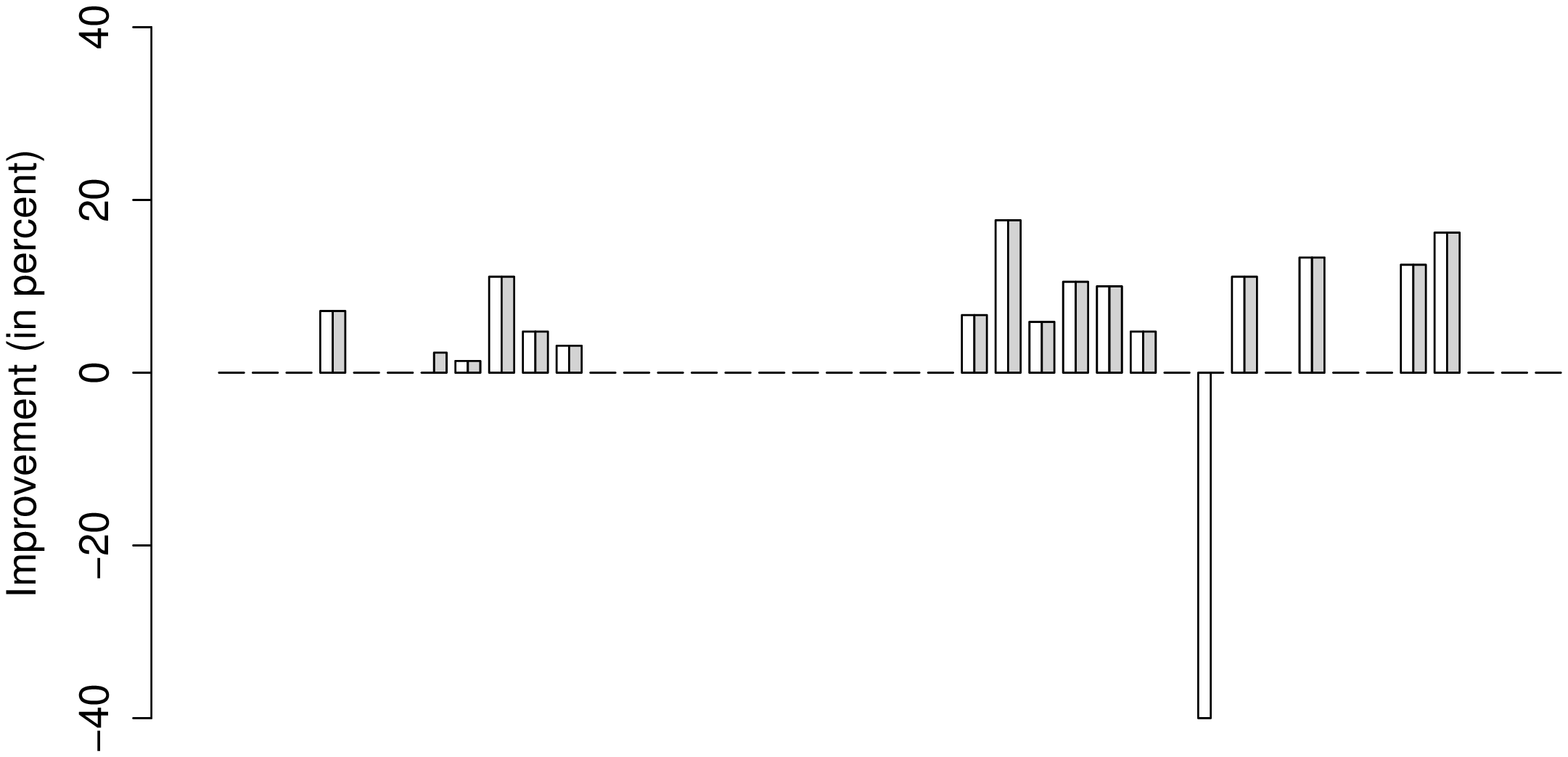}
 \includegraphics[width=5cm,angle=-90]{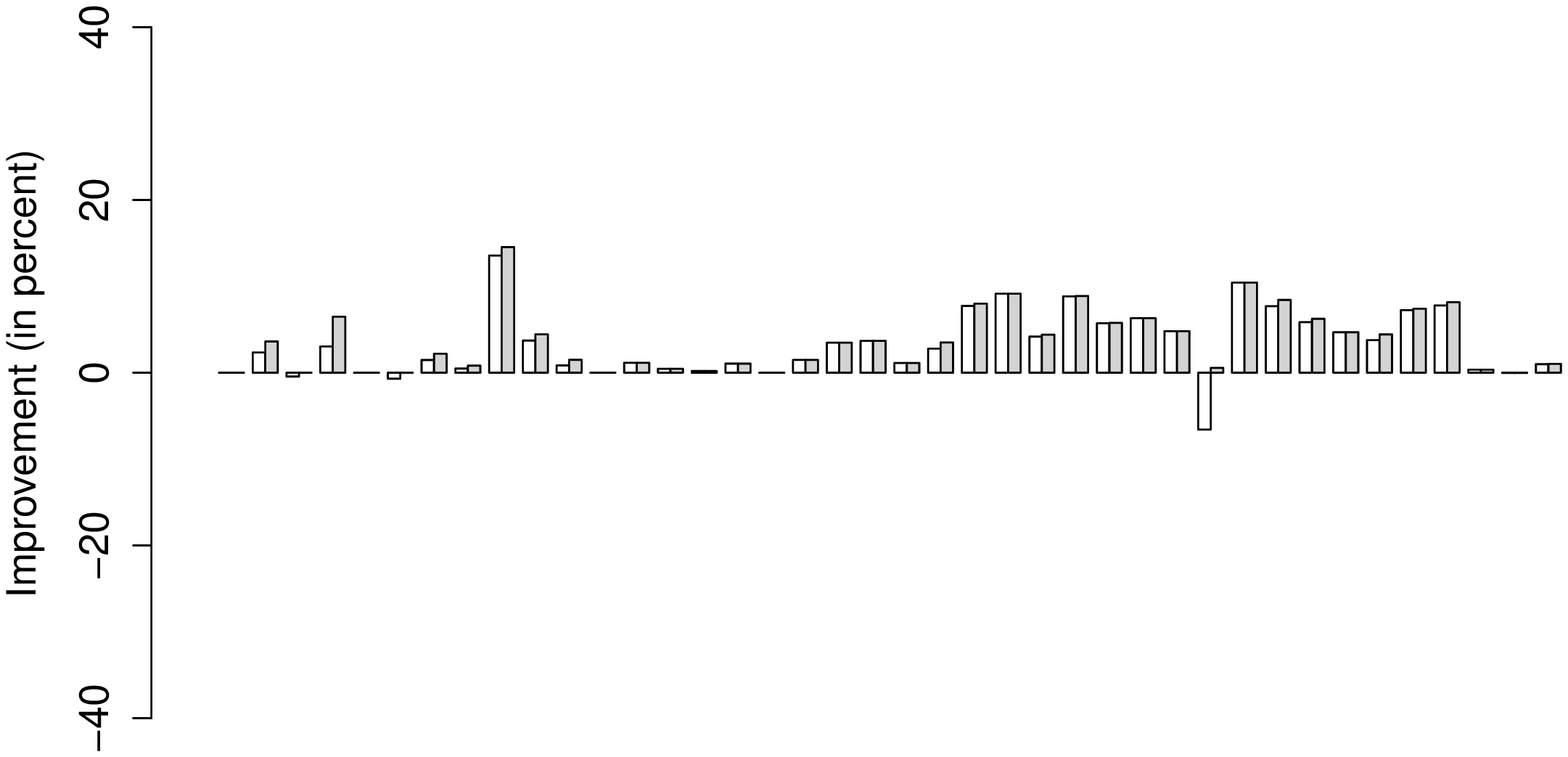}
\caption{Summary of results for the second set of instances from the DIMACS challenge. Both graphics show the performance improvement of \fsim\ (light gray bars) and \fsimp\ (dark gray bars) over \fin\ (in percent). The instances of Table~\ref{tab:res-dimacs2} are treated from left to right in the same order. The top graphic concerns the best colorings found, whereas the bottom graphic concerns the average solution quality.}
\label{fig:res-dimacs2}
\end{figure}

Finally, in Figures~\ref{fig:res-dimacs1} and~\ref{fig:res-dimacs2} the results of Tables~\ref{tab:res-dimacs1} and~\ref{tab:res-dimacs2} are provided again in a graphical form. 

\subsection{Results for Grid Topologies}
\label{sec:grid-topologies}

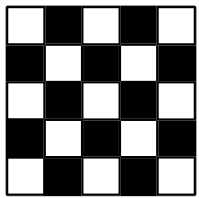
\begin{figure}[!ht]
\centering
 \psset{xunit=.5mm,yunit=.5mm,runit=.5mm}
 \begin{pspicture}(60,60)
 \psline[linewidth=1pt,linecolor=black]{-}(5,5)(55,5)
 \psline[linewidth=1pt,linecolor=black]{-}(5,15)(55,15)
 \psline[linewidth=1pt,linecolor=black]{-}(5,25)(55,25)
 \psline[linewidth=1pt,linecolor=black]{-}(5,35)(55,35)
 \psline[linewidth=1pt,linecolor=black]{-}(5,45)(55,45)
 \psline[linewidth=1pt,linecolor=black]{-}(5,55)(55,55)
 \psline[linewidth=1pt,linecolor=black]{-}(5,5)(5,55)
 \psline[linewidth=1pt,linecolor=black]{-}(15,5)(15,55)
 \psline[linewidth=1pt,linecolor=black]{-}(25,5)(25,55)
 \psline[linewidth=1pt,linecolor=black]{-}(35,5)(35,55)
 \psline[linewidth=1pt,linecolor=black]{-}(45,5)(45,55)
 \psline[linewidth=1pt,linecolor=black]{-}(55,5)(55,55)
 \dotnode[dotstyle=square,fillstyle=solid,fillcolor=black,dotscale=4.6](20,30){1}
 \dotnode[dotstyle=square,fillstyle=solid,fillcolor=black,dotscale=4.6](40,30){2}
 \dotnode[dotstyle=square,fillstyle=solid,fillcolor=white,dotscale=4.6](10,10){3}
 \dotnode[dotstyle=square,fillstyle=solid,fillcolor= black,dotscale=4.6](30,20){4}
 \dotnode[dotstyle=square,fillstyle=solid,fillcolor=black,dotscale=4.6](10,40){5}
 \dotnode[dotstyle=square,fillstyle=solid,fillcolor=black,dotscale=4.6](30,40){6}
 \dotnode[dotstyle=square,fillstyle=solid,fillcolor=black,dotscale=4.6](40,10){7}
 \dotnode[dotstyle=square,fillstyle=solid,fillcolor=black,dotscale=4.6](20,50){8}
 \dotnode[dotstyle=square,fillstyle=solid,fillcolor=black,dotscale=4.6](40,50){9}
 \dotnode[dotstyle=square,fillstyle=solid,fillcolor=black,dotscale=4.6](50,40){10}
 \dotnode[dotstyle=square,fillstyle=solid,fillcolor=black,dotscale=4.6](50,20){11}
 \dotnode[dotstyle=square,fillstyle=solid,fillcolor=white,dotscale=4.6](40,20){12}
 \dotnode[dotstyle=square,fillstyle=solid,fillcolor=white,dotscale=4.6](40,40){13}
 \dotnode[dotstyle=square,fillstyle=solid,fillcolor=white,dotscale=4.6](20,40){14}
 \dotnode[dotstyle=square,fillstyle=solid,fillcolor=black,dotscale=4.6](20,10){15}
 \dotnode[dotstyle=square,fillstyle=solid,fillcolor=white,dotscale=4.6](10,30){16}
 \dotnode[dotstyle=square,fillstyle=solid,fillcolor=white,dotscale=4.6](30,30){17}
 \dotnode[dotstyle=square,fillstyle=solid,fillcolor=white,dotscale=4.6](10,50){18}
 \dotnode[dotstyle=square,fillstyle=solid,fillcolor=white,dotscale=4.6](30,50){19}
 \dotnode[dotstyle=square,fillstyle=solid,fillcolor=white,dotscale=4.6](50,50){20}
 \dotnode[dotstyle=square,fillstyle=solid,fillcolor=white,dotscale=4.6](50,30){21}
 \dotnode[dotstyle=square,fillstyle=solid,fillcolor=white,dotscale=4.6](50,10){22}
 \dotnode[dotstyle=square,fillstyle=solid,fillcolor=black,dotscale=4.6](10,20){23}
 \dotnode[dotstyle=square,fillstyle=solid,fillcolor=white,dotscale=4.6](30,10){24}
 \dotnode[dotstyle=square,fillstyle=solid,fillcolor=white,dotscale=4.6](20,20){25}
 \end{pspicture}
\caption{Optimal coloring of a grid topology}
\label{fig:grid-correctly-colored}
\end{figure}

Grid topologies are frequently used in various application areas of sensor networks. In theory, the coloring of grids is very simple. They all can be painted as a chessboard, requiring only two colors. For an example see Figure~\ref{fig:grid-correctly-colored}. 

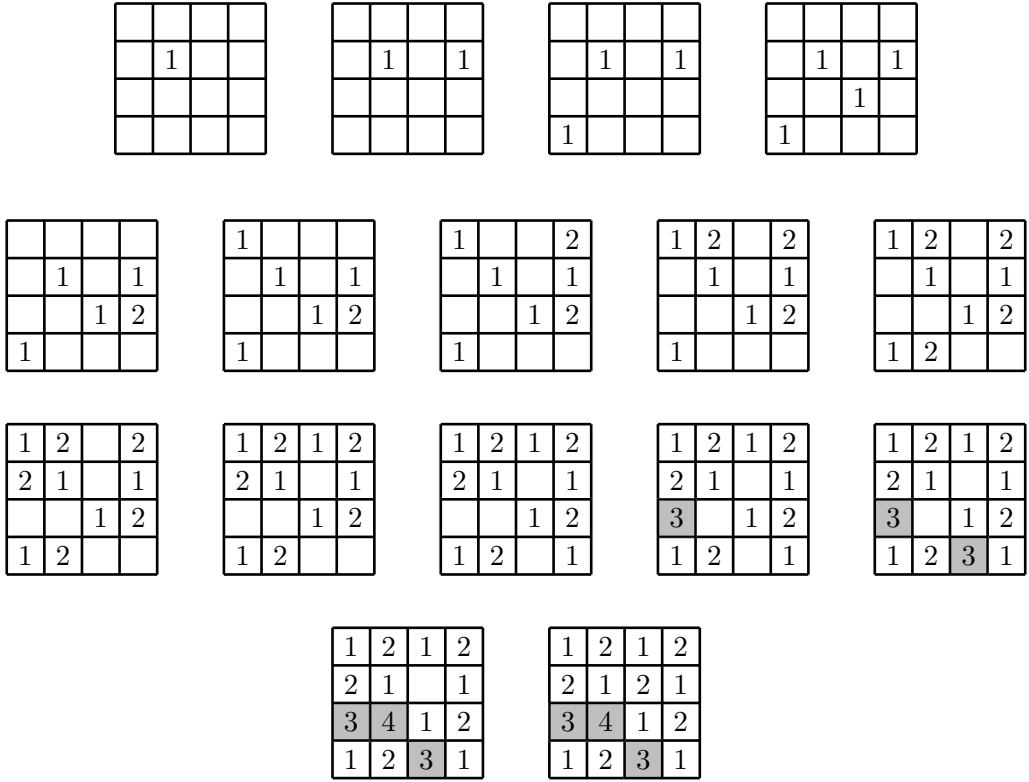
\begin{figure}[!ht]
\centering
\subfigure{
\label{fig:grid-example-state-1}
 \psset{xunit=.5mm,yunit=.5mm,runit=.5mm}
 \begin{pspicture}(50,50)
 \psline[linewidth=1pt,linecolor=black]{-}(5,5)(45,5)
 \psline[linewidth=1pt,linecolor=black]{-}(5,15)(45,15)
 \psline[linewidth=1pt,linecolor=black]{-}(5,25)(45,25)
 \psline[linewidth=1pt,linecolor=black]{-}(5,35)(45,35)
 \psline[linewidth=1pt,linecolor=black]{-}(5,45)(45,45)
% \psline[linewidth=1pt,linecolor=black]{-}(5,55)(55,55)
 \psline[linewidth=1pt,linecolor=black]{-}(5,5)(5,45)
 \psline[linewidth=1pt,linecolor=black]{-}(15,5)(15,45)
 \psline[linewidth=1pt,linecolor=black]{-}(25,5)(25,45)
 \psline[linewidth=1pt,linecolor=black]{-}(35,5)(35,45)
 \psline[linewidth=1pt,linecolor=black]{-}(45,5)(45,45)
% \psline[linewidth=1pt,linecolor=black]{-}(55,5)(55,55)
 %\dotnode[dotstyle=square,fillstyle=solid,fillcolor=blue,dotscale=4.6](20,30){1}
\rput(20,30){\rnode{1}{1}}
 
 \end{pspicture}
}
\subfigure{
\label{fig:grid-example-state-2}
 \psset{xunit=.5mm,yunit=.5mm,runit=.5mm}
 \begin{pspicture}(50,50)
 \psline[linewidth=1pt,linecolor=black]{-}(5,5)(45,5)
 \psline[linewidth=1pt,linecolor=black]{-}(5,15)(45,15)
 \psline[linewidth=1pt,linecolor=black]{-}(5,25)(45,25)
 \psline[linewidth=1pt,linecolor=black]{-}(5,35)(45,35)
 \psline[linewidth=1pt,linecolor=black]{-}(5,45)(45,45)
% \psline[linewidth=1pt,linecolor=black]{-}(5,55)(55,55)
 \psline[linewidth=1pt,linecolor=black]{-}(5,5)(5,45)
 \psline[linewidth=1pt,linecolor=black]{-}(15,5)(15,45)
 \psline[linewidth=1pt,linecolor=black]{-}(25,5)(25,45)
 \psline[linewidth=1pt,linecolor=black]{-}(35,5)(35,45)
 \psline[linewidth=1pt,linecolor=black]{-}(45,5)(45,45)
% \psline[linewidth=1pt,linecolor=black]{-}(55,5)(55,55)

\rput(20,30){\rnode{1}{1}}
\rput(40,30){\rnode{1}{1}}

 \end{pspicture}
}
\subfigure{
\label{fig:grid-example-state-3}
 \psset{xunit=.5mm,yunit=.5mm,runit=.5mm}
 \begin{pspicture}(50,50)
 \psline[linewidth=1pt,linecolor=black]{-}(5,5)(45,5)
 \psline[linewidth=1pt,linecolor=black]{-}(5,15)(45,15)
 \psline[linewidth=1pt,linecolor=black]{-}(5,25)(45,25)
 \psline[linewidth=1pt,linecolor=black]{-}(5,35)(45,35)
 \psline[linewidth=1pt,linecolor=black]{-}(5,45)(45,45)
% \psline[linewidth=1pt,linecolor=black]{-}(5,55)(55,55)
 \psline[linewidth=1pt,linecolor=black]{-}(5,5)(5,45)
 \psline[linewidth=1pt,linecolor=black]{-}(15,5)(15,45)
 \psline[linewidth=1pt,linecolor=black]{-}(25,5)(25,45)
 \psline[linewidth=1pt,linecolor=black]{-}(35,5)(35,45)
 \psline[linewidth=1pt,linecolor=black]{-}(45,5)(45,45)
% \psline[linewidth=1pt,linecolor=black]{-}(55,5)(55,55)

 \rput(20,30){\rnode{1}{1}}
 \rput(40,30){\rnode{2}{1}}
 \rput(10,10){\rnode{3}{1}}
 \end{pspicture}
}
\subfigure{
\label{fig:grid-example-state-4}
 \psset{xunit=.5mm,yunit=.5mm,runit=.5mm}
 \begin{pspicture}(50,50)
 \psline[linewidth=1pt,linecolor=black]{-}(5,5)(45,5)
 \psline[linewidth=1pt,linecolor=black]{-}(5,15)(45,15)
 \psline[linewidth=1pt,linecolor=black]{-}(5,25)(45,25)
 \psline[linewidth=1pt,linecolor=black]{-}(5,35)(45,35)
 \psline[linewidth=1pt,linecolor=black]{-}(5,45)(45,45)
% \psline[linewidth=1pt,linecolor=black]{-}(5,55)(55,55)
 \psline[linewidth=1pt,linecolor=black]{-}(5,5)(5,45)
 \psline[linewidth=1pt,linecolor=black]{-}(15,5)(15,45)
 \psline[linewidth=1pt,linecolor=black]{-}(25,5)(25,45)
 \psline[linewidth=1pt,linecolor=black]{-}(35,5)(35,45)
 \psline[linewidth=1pt,linecolor=black]{-}(45,5)(45,45)
% \psline[linewidth=1pt,linecolor=black]{-}(55,5)(55,55)
 
 \rput(20,30){\rnode{1}{1}}
 \rput(40,30){\rnode{2}{1}}
 \rput(10,10){\rnode{3}{1}}
 \rput(30,20){\rnode{4}{1}}
 
 \end{pspicture}
}

\subfigure{
\label{fig:grid-example-state-5}
 \psset{xunit=.5mm,yunit=.5mm,runit=.5mm}
 \begin{pspicture}(50,50)
 \psline[linewidth=1pt,linecolor=black]{-}(5,5)(45,5)
 \psline[linewidth=1pt,linecolor=black]{-}(5,15)(45,15)
 \psline[linewidth=1pt,linecolor=black]{-}(5,25)(45,25)
 \psline[linewidth=1pt,linecolor=black]{-}(5,35)(45,35)
 \psline[linewidth=1pt,linecolor=black]{-}(5,45)(45,45)
% \psline[linewidth=1pt,linecolor=black]{-}(5,55)(55,55)
 \psline[linewidth=1pt,linecolor=black]{-}(5,5)(5,45)
 \psline[linewidth=1pt,linecolor=black]{-}(15,5)(15,45)
 \psline[linewidth=1pt,linecolor=black]{-}(25,5)(25,45)
 \psline[linewidth=1pt,linecolor=black]{-}(35,5)(35,45)
 \psline[linewidth=1pt,linecolor=black]{-}(45,5)(45,45)
% \psline[linewidth=1pt,linecolor=black]{-}(55,5)(55,55)

 \rput(20,30){\rnode{1}{1}}
 \rput(40,30){\rnode{2}{1}}
 \rput(10,10){\rnode{3}{1}}
 \rput(30,20){\rnode{4}{1}}
 \rput(40,20){\rnode{5}{2}}
 
 \end{pspicture}
}
\subfigure{
\label{fig:grid-example-state-6}
 \psset{xunit=.5mm,yunit=.5mm,runit=.5mm}
 \begin{pspicture}(50,50)
 \psline[linewidth=1pt,linecolor=black]{-}(5,5)(45,5)
 \psline[linewidth=1pt,linecolor=black]{-}(5,15)(45,15)
 \psline[linewidth=1pt,linecolor=black]{-}(5,25)(45,25)
 \psline[linewidth=1pt,linecolor=black]{-}(5,35)(45,35)
 \psline[linewidth=1pt,linecolor=black]{-}(5,45)(45,45)
% \psline[linewidth=1pt,linecolor=black]{-}(5,55)(55,55)
 \psline[linewidth=1pt,linecolor=black]{-}(5,5)(5,45)
 \psline[linewidth=1pt,linecolor=black]{-}(15,5)(15,45)
 \psline[linewidth=1pt,linecolor=black]{-}(25,5)(25,45)
 \psline[linewidth=1pt,linecolor=black]{-}(35,5)(35,45)
 \psline[linewidth=1pt,linecolor=black]{-}(45,5)(45,45)
% \psline[linewidth=1pt,linecolor=black]{-}(55,5)(55,55)
 
  \rput(20,30){\rnode{1}{1}}
 \rput(40,30){\rnode{2}{1}}
 \rput(10,10){\rnode{3}{1}}
 \rput(30,20){\rnode{4}{1}}
 \rput(40,20){\rnode{5}{2}}
 \rput(10,40){\rnode{6}{1}}
 
 \end{pspicture}
}
\subfigure{
\label{fig:grid-example-state-7}
 \psset{xunit=.5mm,yunit=.5mm,runit=.5mm}
 \begin{pspicture}(50,50)
 \psline[linewidth=1pt,linecolor=black]{-}(5,5)(45,5)
 \psline[linewidth=1pt,linecolor=black]{-}(5,15)(45,15)
 \psline[linewidth=1pt,linecolor=black]{-}(5,25)(45,25)
 \psline[linewidth=1pt,linecolor=black]{-}(5,35)(45,35)
 \psline[linewidth=1pt,linecolor=black]{-}(5,45)(45,45)
% \psline[linewidth=1pt,linecolor=black]{-}(5,55)(55,55)
 \psline[linewidth=1pt,linecolor=black]{-}(5,5)(5,45)
 \psline[linewidth=1pt,linecolor=black]{-}(15,5)(15,45)
 \psline[linewidth=1pt,linecolor=black]{-}(25,5)(25,45)
 \psline[linewidth=1pt,linecolor=black]{-}(35,5)(35,45)
 \psline[linewidth=1pt,linecolor=black]{-}(45,5)(45,45)
% \psline[linewidth=1pt,linecolor=black]{-}(55,5)(55,55)

 \rput(20,30){\rnode{1}{1}}
 \rput(40,30){\rnode{2}{1}}
 \rput(10,10){\rnode{3}{1}}
 \rput(30,20){\rnode{4}{1}}
 \rput(40,20){\rnode{5}{2}}
 \rput(10,40){\rnode{6}{1}}
 \rput(40,40){\rnode{7}{2}}
 
 \end{pspicture}
}
\subfigure{
\label{fig:grid-example-state-8}
 \psset{xunit=.5mm,yunit=.5mm,runit=.5mm}
 \begin{pspicture}(50,50)
 \psline[linewidth=1pt,linecolor=black]{-}(5,5)(45,5)
 \psline[linewidth=1pt,linecolor=black]{-}(5,15)(45,15)
 \psline[linewidth=1pt,linecolor=black]{-}(5,25)(45,25)
 \psline[linewidth=1pt,linecolor=black]{-}(5,35)(45,35)
 \psline[linewidth=1pt,linecolor=black]{-}(5,45)(45,45)
% \psline[linewidth=1pt,linecolor=black]{-}(5,55)(55,55)
 \psline[linewidth=1pt,linecolor=black]{-}(5,5)(5,45)
 \psline[linewidth=1pt,linecolor=black]{-}(15,5)(15,45)
 \psline[linewidth=1pt,linecolor=black]{-}(25,5)(25,45)
 \psline[linewidth=1pt,linecolor=black]{-}(35,5)(35,45)
 \psline[linewidth=1pt,linecolor=black]{-}(45,5)(45,45)
% \psline[linewidth=1pt,linecolor=black]{-}(55,5)(55,55)
 
  \rput(20,30){\rnode{1}{1}}
 \rput(40,30){\rnode{2}{1}}
 \rput(10,10){\rnode{3}{1}}
 \rput(30,20){\rnode{4}{1}}
 \rput(40,20){\rnode{5}{2}}
 \rput(10,40){\rnode{6}{1}}
 \rput(40,40){\rnode{7}{2}}
 \rput(20,40){\rnode{8}{2}}
 
 \end{pspicture}
}
\subfigure{
\label{fig:grid-example-state-9}
 \psset{xunit=.5mm,yunit=.5mm,runit=.5mm}
 \begin{pspicture}(50,50)
 \psline[linewidth=1pt,linecolor=black]{-}(5,5)(45,5)
 \psline[linewidth=1pt,linecolor=black]{-}(5,15)(45,15)
 \psline[linewidth=1pt,linecolor=black]{-}(5,25)(45,25)
 \psline[linewidth=1pt,linecolor=black]{-}(5,35)(45,35)
 \psline[linewidth=1pt,linecolor=black]{-}(5,45)(45,45)
% \psline[linewidth=1pt,linecolor=black]{-}(5,55)(55,55)
 \psline[linewidth=1pt,linecolor=black]{-}(5,5)(5,45)
 \psline[linewidth=1pt,linecolor=black]{-}(15,5)(15,45)
 \psline[linewidth=1pt,linecolor=black]{-}(25,5)(25,45)
 \psline[linewidth=1pt,linecolor=black]{-}(35,5)(35,45)
 \psline[linewidth=1pt,linecolor=black]{-}(45,5)(45,45)
% \psline[linewidth=1pt,linecolor=black]{-}(55,5)(55,55)
 
  \rput(20,30){\rnode{1}{1}}
 \rput(40,30){\rnode{2}{1}}
 \rput(10,10){\rnode{3}{1}}
 \rput(30,20){\rnode{4}{1}}
 \rput(40,20){\rnode{5}{2}}
 \rput(10,40){\rnode{6}{1}}
 \rput(40,40){\rnode{7}{2}}
 \rput(20,40){\rnode{8}{2}}
 \rput(20,10){\rnode{9}{2}}
 
 \end{pspicture}
}
\subfigure{
\label{fig:grid-example-state-10}
 \psset{xunit=.5mm,yunit=.5mm,runit=.5mm}
 \begin{pspicture}(50,50)
 \psline[linewidth=1pt,linecolor=black]{-}(5,5)(45,5)
 \psline[linewidth=1pt,linecolor=black]{-}(5,15)(45,15)
 \psline[linewidth=1pt,linecolor=black]{-}(5,25)(45,25)
 \psline[linewidth=1pt,linecolor=black]{-}(5,35)(45,35)
 \psline[linewidth=1pt,linecolor=black]{-}(5,45)(45,45)
% \psline[linewidth=1pt,linecolor=black]{-}(5,55)(55,55)
 \psline[linewidth=1pt,linecolor=black]{-}(5,5)(5,45)
 \psline[linewidth=1pt,linecolor=black]{-}(15,5)(15,45)
 \psline[linewidth=1pt,linecolor=black]{-}(25,5)(25,45)
 \psline[linewidth=1pt,linecolor=black]{-}(35,5)(35,45)
 \psline[linewidth=1pt,linecolor=black]{-}(45,5)(45,45)
% \psline[linewidth=1pt,linecolor=black]{-}(55,5)(55,55)

 \rput(20,30){\rnode{1}{1}}
 \rput(40,30){\rnode{2}{1}}
 \rput(10,10){\rnode{3}{1}}
 \rput(30,20){\rnode{4}{1}}
 \rput(40,20){\rnode{5}{2}}
 \rput(10,40){\rnode{6}{1}}
 \rput(40,40){\rnode{7}{2}}
 \rput(20,40){\rnode{8}{2}}
 \rput(20,10){\rnode{9}{2}}
 \rput(10,30){\rnode{10}{2}}

 \end{pspicture}
}
\subfigure{
\label{fig:grid-example-state-11}
 \psset{xunit=.5mm,yunit=.5mm,runit=.5mm}
 \begin{pspicture}(50,50)
 \psline[linewidth=1pt,linecolor=black]{-}(5,5)(45,5)
 \psline[linewidth=1pt,linecolor=black]{-}(5,15)(45,15)
 \psline[linewidth=1pt,linecolor=black]{-}(5,25)(45,25)
 \psline[linewidth=1pt,linecolor=black]{-}(5,35)(45,35)
 \psline[linewidth=1pt,linecolor=black]{-}(5,45)(45,45)
% \psline[linewidth=1pt,linecolor=black]{-}(5,55)(55,55)
 \psline[linewidth=1pt,linecolor=black]{-}(5,5)(5,45)
 \psline[linewidth=1pt,linecolor=black]{-}(15,5)(15,45)
 \psline[linewidth=1pt,linecolor=black]{-}(25,5)(25,45)
 \psline[linewidth=1pt,linecolor=black]{-}(35,5)(35,45)
 \psline[linewidth=1pt,linecolor=black]{-}(45,5)(45,45)
% \psline[linewidth=1pt,linecolor=black]{-}(55,5)(55,55)
 
 \rput(20,30){\rnode{1}{1}}
 \rput(40,30){\rnode{2}{1}}
 \rput(10,10){\rnode{3}{1}}
 \rput(30,20){\rnode{4}{1}}
 \rput(40,20){\rnode{5}{2}}
 \rput(10,40){\rnode{6}{1}}
 \rput(40,40){\rnode{7}{2}}
 \rput(20,40){\rnode{8}{2}}
 \rput(20,10){\rnode{9}{2}}
 \rput(10,30){\rnode{10}{2}}
 \rput(30,40){\rnode{11}{1}}
 
  \end{pspicture}
}
\subfigure{
\label{fig:grid-example-state-12}
 \psset{xunit=.5mm,yunit=.5mm,runit=.5mm}
 \begin{pspicture}(50,50)
 \psline[linewidth=1pt,linecolor=black]{-}(5,5)(45,5)
 \psline[linewidth=1pt,linecolor=black]{-}(5,15)(45,15)
 \psline[linewidth=1pt,linecolor=black]{-}(5,25)(45,25)
 \psline[linewidth=1pt,linecolor=black]{-}(5,35)(45,35)
 \psline[linewidth=1pt,linecolor=black]{-}(5,45)(45,45)
% \psline[linewidth=1pt,linecolor=black]{-}(5,55)(55,55)
 \psline[linewidth=1pt,linecolor=black]{-}(5,5)(5,45)
 \psline[linewidth=1pt,linecolor=black]{-}(15,5)(15,45)
 \psline[linewidth=1pt,linecolor=black]{-}(25,5)(25,45)
 \psline[linewidth=1pt,linecolor=black]{-}(35,5)(35,45)
 \psline[linewidth=1pt,linecolor=black]{-}(45,5)(45,45)
% \psline[linewidth=1pt,linecolor=black]{-}(55,5)(55,55)
 
  \rput(20,30){\rnode{1}{1}}
 \rput(40,30){\rnode{2}{1}}
 \rput(10,10){\rnode{3}{1}}
 \rput(30,20){\rnode{4}{1}}
 \rput(40,20){\rnode{5}{2}}
 \rput(10,40){\rnode{6}{1}}
 \rput(40,40){\rnode{7}{2}}
 \rput(20,40){\rnode{8}{2}}
 \rput(20,10){\rnode{9}{2}}
 \rput(10,30){\rnode{10}{2}}
 \rput(30,40){\rnode{11}{1}}
 \rput(40,10){\rnode{12}{1}}
 
 \end{pspicture}
}
\subfigure{
\label{fig:grid-example-state-13}
 \psset{xunit=.5mm,yunit=.5mm,runit=.5mm}
 \begin{pspicture}(50,50)
 \psline[linewidth=1pt,linecolor=black]{-}(5,5)(45,5)
 \psline[linewidth=1pt,linecolor=black]{-}(5,15)(45,15)
 \psline[linewidth=1pt,linecolor=black]{-}(5,25)(45,25)
 \psline[linewidth=1pt,linecolor=black]{-}(5,35)(45,35)
 \psline[linewidth=1pt,linecolor=black]{-}(5,45)(45,45)
% \psline[linewidth=1pt,linecolor=black]{-}(5,55)(55,55)
 \psline[linewidth=1pt,linecolor=black]{-}(5,5)(5,45)
 \psline[linewidth=1pt,linecolor=black]{-}(15,5)(15,45)
 \psline[linewidth=1pt,linecolor=black]{-}(25,5)(25,45)
 \psline[linewidth=1pt,linecolor=black]{-}(35,5)(35,45)
 \psline[linewidth=1pt,linecolor=black]{-}(45,5)(45,45)
% \psline[linewidth=1pt,linecolor=black]{-}(55,5)(55,55)

\dotnode[dotstyle=square,fillstyle=solid,fillcolor=lightgray,dotscale=4.6](10,20){b1}

 \rput(20,30){\rnode{1}{1}}
 \rput(40,30){\rnode{2}{1}}
 \rput(10,10){\rnode{3}{1}}
 \rput(30,20){\rnode{4}{1}}
 \rput(40,20){\rnode{5}{2}}
 \rput(10,40){\rnode{6}{1}}
 \rput(40,40){\rnode{7}{2}}
 \rput(20,40){\rnode{8}{2}}
 \rput(20,10){\rnode{9}{2}}
 \rput(10,30){\rnode{10}{2}}
 \rput(30,40){\rnode{11}{1}}
 \rput(40,10){\rnode{12}{1}}
 \rput(10,20){\rnode{13}{3}}

 \end{pspicture}
}
\subfigure{
\label{fig:grid-example-state-14}
 \psset{xunit=.5mm,yunit=.5mm,runit=.5mm}
 \begin{pspicture}(50,50)
 \psline[linewidth=1pt,linecolor=black]{-}(5,5)(45,5)
 \psline[linewidth=1pt,linecolor=black]{-}(5,15)(45,15)
 \psline[linewidth=1pt,linecolor=black]{-}(5,25)(45,25)
 \psline[linewidth=1pt,linecolor=black]{-}(5,35)(45,35)
 \psline[linewidth=1pt,linecolor=black]{-}(5,45)(45,45)
% \psline[linewidth=1pt,linecolor=black]{-}(5,55)(55,55)
 \psline[linewidth=1pt,linecolor=black]{-}(5,5)(5,45)
 \psline[linewidth=1pt,linecolor=black]{-}(15,5)(15,45)
 \psline[linewidth=1pt,linecolor=black]{-}(25,5)(25,45)
 \psline[linewidth=1pt,linecolor=black]{-}(35,5)(35,45)
 \psline[linewidth=1pt,linecolor=black]{-}(45,5)(45,45)
% \psline[linewidth=1pt,linecolor=black]{-}(55,5)(55,55)

\dotnode[dotstyle=square,fillstyle=solid,fillcolor=lightgray,dotscale=4.6](10,20){b1}
\dotnode[dotstyle=square,fillstyle=solid,fillcolor=lightgray,dotscale=4.6](30,10){b2}

 \rput(20,30){\rnode{1}{1}}
 \rput(40,30){\rnode{2}{1}}
 \rput(10,10){\rnode{3}{1}}
 \rput(30,20){\rnode{4}{1}}
 \rput(40,20){\rnode{5}{2}}
 \rput(10,40){\rnode{6}{1}}
 \rput(40,40){\rnode{7}{2}}
 \rput(20,40){\rnode{8}{2}}
 \rput(20,10){\rnode{9}{2}}
 \rput(10,30){\rnode{10}{2}}
 \rput(30,40){\rnode{11}{1}}
 \rput(40,10){\rnode{12}{1}}
 \rput(10,20){\rnode{13}{3}}
 \rput(30,10){\rnode{14}{3}}

 \end{pspicture}
}
\subfigure{
\label{fig:grid-example-state-15}
 \psset{xunit=.5mm,yunit=.5mm,runit=.5mm}
 \begin{pspicture}(50,50)
 \psline[linewidth=1pt,linecolor=black]{-}(5,5)(45,5)
 \psline[linewidth=1pt,linecolor=black]{-}(5,15)(45,15)
 \psline[linewidth=1pt,linecolor=black]{-}(5,25)(45,25)
 \psline[linewidth=1pt,linecolor=black]{-}(5,35)(45,35)
 \psline[linewidth=1pt,linecolor=black]{-}(5,45)(45,45)
% \psline[linewidth=1pt,linecolor=black]{-}(5,55)(55,55)
 \psline[linewidth=1pt,linecolor=black]{-}(5,5)(5,45)
 \psline[linewidth=1pt,linecolor=black]{-}(15,5)(15,45)
 \psline[linewidth=1pt,linecolor=black]{-}(25,5)(25,45)
 \psline[linewidth=1pt,linecolor=black]{-}(35,5)(35,45)
 \psline[linewidth=1pt,linecolor=black]{-}(45,5)(45,45)
% \psline[linewidth=1pt,linecolor=black]{-}(55,5)(55,55)
 
\dotnode[dotstyle=square,fillstyle=solid,fillcolor=lightgray,dotscale=4.6](10,20){b1}
\dotnode[dotstyle=square,fillstyle=solid,fillcolor=lightgray,dotscale=4.6](30,10){b2}
\dotnode[dotstyle=square,fillstyle=solid,fillcolor=lightgray,dotscale=4.6](20,20){b3}
 
 \rput(20,30){\rnode{1}{1}}
 \rput(40,30){\rnode{2}{1}}
 \rput(10,10){\rnode{3}{1}}
 \rput(30,20){\rnode{4}{1}}
 \rput(40,20){\rnode{5}{2}}
 \rput(10,40){\rnode{6}{1}}
 \rput(40,40){\rnode{7}{2}}
 \rput(20,40){\rnode{8}{2}}
 \rput(20,10){\rnode{9}{2}}
 \rput(10,30){\rnode{10}{2}}
 \rput(30,40){\rnode{11}{1}}
 \rput(40,10){\rnode{12}{1}}
 \rput(10,20){\rnode{13}{3}}
 \rput(30,10){\rnode{14}{3}}
 \rput(20,20){\rnode{15}{4}}
 \end{pspicture}
}
\subfigure{
\label{fig:grid-example-state-16}
 \psset{xunit=.5mm,yunit=.5mm,runit=.5mm}
 \begin{pspicture}(50,50)
 \psline[linewidth=1pt,linecolor=black]{-}(5,5)(45,5)
 \psline[linewidth=1pt,linecolor=black]{-}(5,15)(45,15)
 \psline[linewidth=1pt,linecolor=black]{-}(5,25)(45,25)
 \psline[linewidth=1pt,linecolor=black]{-}(5,35)(45,35)
 \psline[linewidth=1pt,linecolor=black]{-}(5,45)(45,45)
% \psline[linewidth=1pt,linecolor=black]{-}(5,55)(55,55)
 \psline[linewidth=1pt,linecolor=black]{-}(5,5)(5,45)
 \psline[linewidth=1pt,linecolor=black]{-}(15,5)(15,45)
 \psline[linewidth=1pt,linecolor=black]{-}(25,5)(25,45)
 \psline[linewidth=1pt,linecolor=black]{-}(35,5)(35,45)
 \psline[linewidth=1pt,linecolor=black]{-}(45,5)(45,45)
% \psline[linewidth=1pt,linecolor=black]{-}(55,5)(55,55)
 
% \dotnode[dotstyle=square,fillstyle=solid,fillcolor=blue,dotscale=4.6](20,30){1}
% \dotnode[dotstyle=square,fillstyle=solid,fillcolor=blue,dotscale=4.6](40,30){2}
% \dotnode[dotstyle=square,fillstyle=solid,fillcolor=blue,dotscale=4.6](10,10){3}
% \dotnode[dotstyle=square,fillstyle=solid,fillcolor= blue,dotscale=4.6](30,20){4}
% \dotnode[dotstyle=square,fillstyle=solid,fillcolor=red,dotscale=4.6](40,20){5}
% \dotnode[dotstyle=square,fillstyle=solid,fillcolor=blue,dotscale=4.6](10,40){6}
% \dotnode[dotstyle=square,fillstyle=solid,fillcolor=red,dotscale=4.6](40,40){7}
% \dotnode[dotstyle=square,fillstyle=solid,fillcolor=red,dotscale=4.6](20,40){8}
% \dotnode[dotstyle=square,fillstyle=solid,fillcolor=red,dotscale=4.6](20,10){9}
% \dotnode[dotstyle=square,fillstyle=solid,fillcolor=red,dotscale=4.6](10,30){10}
% \dotnode[dotstyle=square,fillstyle=solid,fillcolor=blue,dotscale=4.6](30,40){11}
% \dotnode[dotstyle=square,fillstyle=solid,fillcolor=blue,dotscale=4.6](40,10){12}
% \dotnode[dotstyle=square,fillstyle=solid,fillcolor=green,dotscale=4.6](10,20){13}
% \dotnode[dotstyle=square,fillstyle=solid,fillcolor=green,dotscale=4.6](30,10){14}
% \dotnode[dotstyle=square,fillstyle=solid,fillcolor=yellow,dotscale=4.6](20,20){15}
% \dotnode[dotstyle=square,fillstyle=solid,fillcolor=red,dotscale=4.6](30,30){16}

\dotnode[dotstyle=square,fillstyle=solid,fillcolor=lightgray,dotscale=4.6](10,20){b1}
\dotnode[dotstyle=square,fillstyle=solid,fillcolor=lightgray,dotscale=4.6](30,10){b2}
\dotnode[dotstyle=square,fillstyle=solid,fillcolor=lightgray,dotscale=4.6](20,20){b3}

 \rput(20,30){\rnode{1}{1}}
 \rput(40,30){\rnode{2}{1}}
 \rput(10,10){\rnode{3}{1}}
 \rput(30,20){\rnode{4}{1}}
 \rput(40,20){\rnode{5}{2}}
 \rput(10,40){\rnode{6}{1}}
 \rput(40,40){\rnode{7}{2}}
 \rput(20,40){\rnode{8}{2}}
 \rput(20,10){\rnode{9}{2}}
 \rput(10,30){\rnode{10}{2}}
 \rput(30,40){\rnode{11}{1}}
 \rput(40,10){\rnode{12}{1}} \rput(10,20){\rnode{13}{3}}
 \rput(30,10){\rnode{14}{3}}
 \rput(20,20){\rnode{15}{4}}
 \rput(30,30){\rnode{16}{2}}
 \end{pspicture}
}
\caption{Example of the distributed incremental process of coloring a grid of $4\times 4$ nodes. The nodes choose colors in a certain order, one node at a time. The process starts at the top left grid and ends at the bottom right grid. Due to early decisions (see the top row), the last row shows the formation of borders (see the gray-colored nodes) where additional colors are needed for achieving valid colorings.}
\label{fig:grid-example-state}
\end{figure}

The way in which an optimal coloring can easily be achieved is to start the coloring process in a single node with the first color, and then proceed incrementally. The next step consists in coloring all the neighbors of the starting node in the second color. All the neighbors of these nodes have to be painted in the first color again, and so on. However, when considering distributed computing, nodes only have local information, whereas information about the position in the grid is missing. Moreover, the incremental process described above is difficult to achieve without a global control. Therefore, when coloring grids in a distributed way, what usually happens is that the coloring process is initiated in several different nodes. If the coloring of these nodes does not follow the chessboard distribution of colors, eventually borders will form where additional colors are needed in order to obtain valid colorings. An example is shown in Figure~\ref{fig:grid-example-state}. In this context, remember that numbers correspond to colors. The process of an incremental coloring is shown starting at the top left grid and ending at the bottom right grid. The first row shows several nodes where the coloring is initiated with color 1. These wrong initial decisions lead to borders (see the gray-colored nodes in the bottom row) where additional colors are needed.

\begin{table*}[!t]
\caption{Results for grid (respectivly, torus) topologies.}\label{tab:res-grid}
\centering
\scalebox{0.7}{
\begin{tabular}{ccccrrcrrcrr}
\hline \hline
\multirow{2}{*}{Instance} & $\;\;$ & \multirow{2}{*}{$(n,\Delta,\chi)$} & $\;\;$ & \multicolumn{2}{c}{Finocchi}& $\;\;$ & \multicolumn{2}{c}{$\text{FrogSim}^{\varominus}$} & $\;\;$ &\multicolumn{2}{c}{$\text{FrogSim}$}\\
&& && colors & avg. && colors & avg. && colors & avg. \\
\cline{1-1} \cline{3-3} \cline{5-6} \cline{8-9} \cline{11-12}
grid2x1&&(2,1,2) && {\bf 2}  & 2.000 && {\bf 2}  & 2.000 && {\bf 2}  & 2.000\\
grid2x2&&(4,2,2) && {\bf 2}  & 2.000 && {\bf 2}  & 2.000 && {\bf 2}  & 2.000\\
grid3x1&&(3,2,2) && {\bf 2}  & 2.000 && {\bf 2}  & 2.000 && {\bf 2}  & 2.000\\
grid3x2&&(6,3,2) && 2 & 2.140 && {\bf 2}  & 2.000 && {\bf 2}  & 2.000\\
grid3x3&&(9,4,2) && 2 & 2.360 && 2 & 2.376 && {\bf 2}  & 2.000\\
grid4x1&&(4,2,2) && 2 & 2.250 && {\bf 2}  & 2.000 && {\bf 2}  & 2.000\\
grid4x2&&(8,3,2) && 2 & 2.570 && {\bf 2}  & 2.000 && {\bf 2}  & 2.000\\
grid4x3&&(12,4,2) && 2 & 3.280 && 2 & 2.465 && {\bf 2}  & 2.000\\
grid4x4&&(16,4,2) && 2 & 3.180 && 2 & 2.465 && {\bf 2}  & 2.000\\
grid5x1&&(5,2,2) && 2 & 2.450 && {\bf 2}  & 2.000 && {\bf 2}  & 2.000\\
grid5x2&&(10,3,2) && 2 & 2.600 && 2 & 2.238 && {\bf 2}  & 2.000\\
grid5x3&&(15,4,2) && 2 & 2.420 && 2 & 2.238 && {\bf 2}  & 2.000\\
grid5x4&&(20,4,2) && 2 & 3.350 && 2 & 2.515 && {\bf 2}  & 2.000\\
grid5x5&&(25,4,2) && 2 & 3.470 && 2 & 2.683 && {\bf 2}  & 2.000\\
grid6x1&&(6,2,2) && 2 & 2.870 && {\bf 2}  & 2.000 && {\bf 2}  & 2.000\\
grid6x2&&(12,3,2) && 2 & 2.740 && 2 & 2.535 && {\bf 2}  & 2.000\\
grid6x3&&(18,4,2) && 2 & 3.230 && 2 & 2.426 && {\bf 2}  & 2.000\\
grid6x4&&(24,4,2) && 3 & 3.050 && 2 & 2.980 && {\bf 2}  & 2.000\\
grid6x5&&(30,4,2) && 4 & 4.000 && 2 & 2.931 && {\bf 2}  & 2.000\\
grid6x6&&(36,4,2) && 3 & 3.860 && 2 & 3.069 && {\bf 2}  & 2.000\\
grid7x1&&(7,2,2) && 2 & 2.300 && {\bf 2}  & 2.000 && {\bf 2}  & 2.000\\
grid7x2&&(14,3,2) && 3 & 3.260 && 2 & 2.455 && {\bf 2}  & 2.000\\
grid7x3&&(21,4,2) && 3 & 3.530 && 2 & 2.584 && {\bf 2}  & 2.000\\
grid7x4&&(28,4,2) && 3 & 3.750 && 2 & 3.050 && {\bf 2}  & 2.000\\
grid7x5&&(35,4,2) && 4 & 4.000 && 2 & 3.366 && {\bf 2}  & 2.000\\
grid7x6&&(42,4,2) && 4 & 4.230 && 3 & 3.851 && {\bf 2}  & 2.000\\
grid7x7&&(49,4,2) && 3 & 3.930 && 4 & 4.000 && {\bf 2}  & 2.000\\
grid8x1&&(8,2,2) && 2 & 2.500 && {\bf 2}  & 2.000 && {\bf 2}  & 2.000\\
grid8x2&&(16,3,2) && 2 & 2.620 && 2 & 2.158 && {\bf 2}  & 2.000\\
grid8x3&&(24,4,2) && 3 & 3.730 && 2 & 3.168 && {\bf 2}  & 2.000\\
grid8x4&&(32,4,2) && 2 & 3.570 && 2 & 3.465 && {\bf 2}  & 2.000\\
grid8x5&&(40,4,2) && 3 & 3.800 && 2 & 3.356 && {\bf 2}  & 2.000\\
grid8x6&&(48,4,2) && 4 & 4.000 && 3 & 3.673 && {\bf 2}  & 2.000\\
grid8x7&&(56,4,2) && 4 & 4.000 && 2 & 3.396 && {\bf 2}  & 2.000\\
grid8x8&&(64,4,2) && 4 & 4.130 && 3 & 3.782 && {\bf 2}  & 2.000\\
grid9x1&&(9,2,2) && 2 & 2.630 && 2 & 2.396 && {\bf 2}  & 2.000\\
grid9x2&&(18,3,2) && 2 & 3.590 && 2 & 2.485 && {\bf 2}  & 2.000\\
grid9x3&&(27,4,2) && 3 & 3.860 && 2 & 3.030 && {\bf 2}  & 2.000\\
grid9x4&&(36,4,2) && 4 & 4.000 && 2 & 3.149 && {\bf 2}  & 2.000\\
grid9x5&&(45,4,2) && 4 & 4.000 && 3 & 3.465 && {\bf 2}  & 2.000\\
grid9x6&&(54,4,2) && 4 & 4.010 && 2 & 3.307 && {\bf 2}  & 2.000\\
grid9x7&&(63,4,2) && 4 & 4.000 && 3 & 3.822 && {\bf 2}  & 2.000\\
grid9x8&&(72,4,2) && 4 & 4.000 && 3 & 3.901 && {\bf 2}  & 2.000\\
grid9x9&&(81,4,2) && 4 & 4.000 && 3 & 3.762 && {\bf 2}  & 2.000\\
Ising32x8.col&&(256,4,2) && 4 & 4.000 && 4 & 4.000 && {\bf 2}  & 2.000\\
Ising32x8-torus.col&&(256,4,2) && 4 & 4.000 && 4 & 4.000 && {\bf 2}  & 2.000\\
\cline{1-1} \cline{3-3} \cline{5-6} \cline{8-9} \cline{11-12}
average && && 2.804 & 3.288 && 2.283 & 2.838 && 2.000 & 2.000\\
\cline{1-1} \cline{3-3} \cline{5-6} \cline{8-9} \cline{11-12}
\# times better && && \multicolumn{2}{c}{0} && \multicolumn{2}{c}{0} && \multicolumn{2}{c}{36}\\
\# times all equal && &&\multicolumn{2}{c}{3} && \multicolumn{2}{c}{3} && \multicolumn{2}{c}{3}\\
\# times worse && &&\multicolumn{2}{c}{39} && \multicolumn{2}{c}{2} && \multicolumn{2}{c}{0}\\
\hline \hline
\end{tabular}}
\end{table*}

Computational results are shown in Table~\ref{tab:res-grid}. Note that in this case all chromatic numbers are known as they can be established theoretically. While small grids can basically be colored correctly by all three algorithms, both \fin\ and \fsim\ have---as expected---increasing difficulties when the grid size grows. Although this is the case, \fsim\ has clear advantages over \fin. This is indicated by the average numbers given in the fourth but last table row, and also by the fact that \fin\ is the sole looser in 39 cases, whereas \fsim\ is the sole looser in only 2 cases. In contrast to the deteriorating performance of \fin\ and \fsim\ when the grid size grows, \fsimp\ achieves perfect colorings in all 100 applications for all instances, which is a remarkable achievement. Even the large grids with periodic boundary conditions (see graphs Ising32x8.col and Ising32x8-torus.col used in~\cite{lee2008firefly}) do not pose any difficulty for \fsimp. In contrast, both \fin\ and \fsim\ use four colors instead of the optimal two colors, in each coloring generated. Summarizing we can state that phase~II of FrogSim is very useful when applied to grid topologies, helping the algorithm to achieve an excellent performance. 

\begin{figure}[!t]
\centering
 \includegraphics[width=5cm,angle=-90]{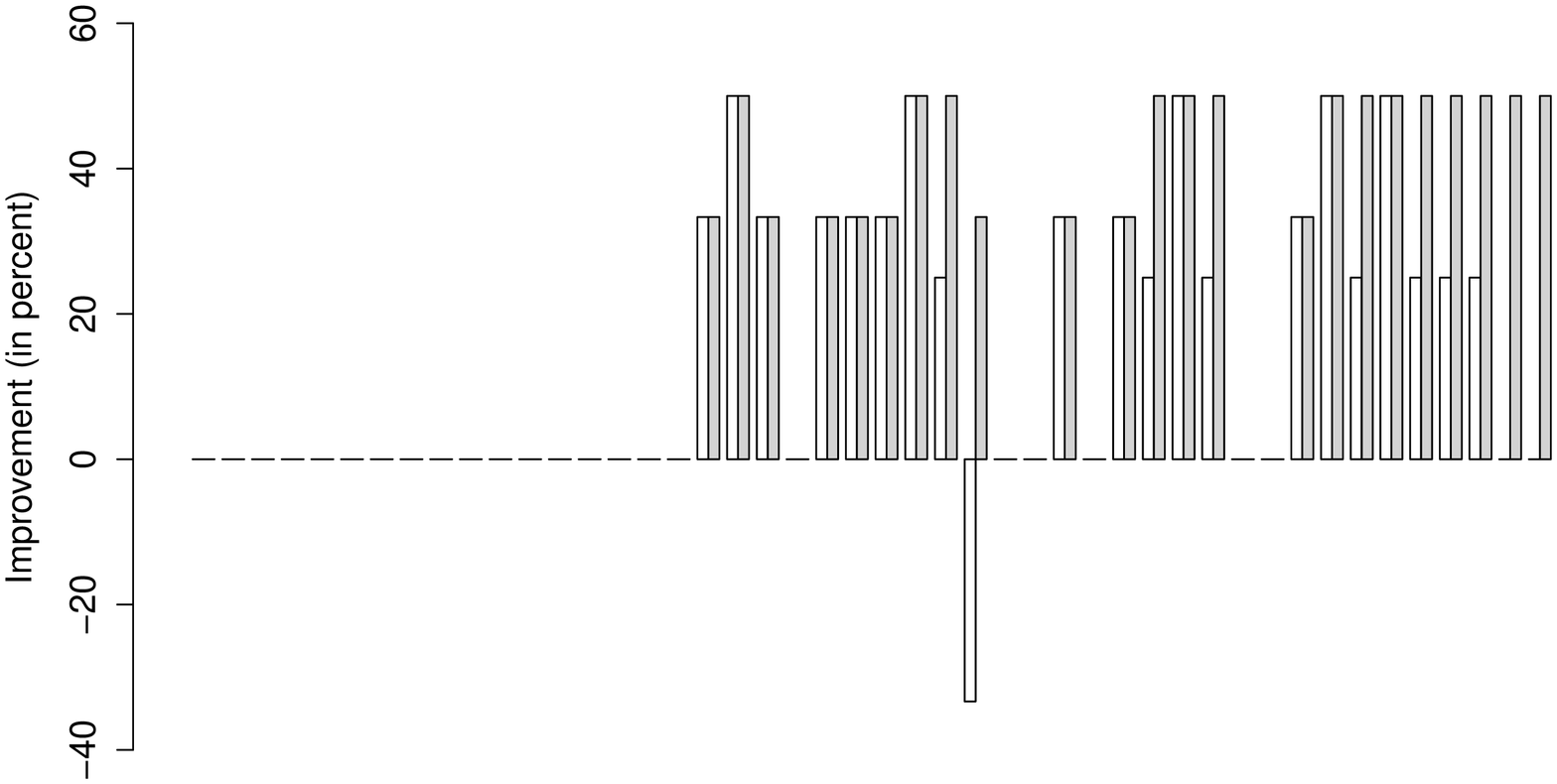}
 \includegraphics[width=5cm,angle=-90]{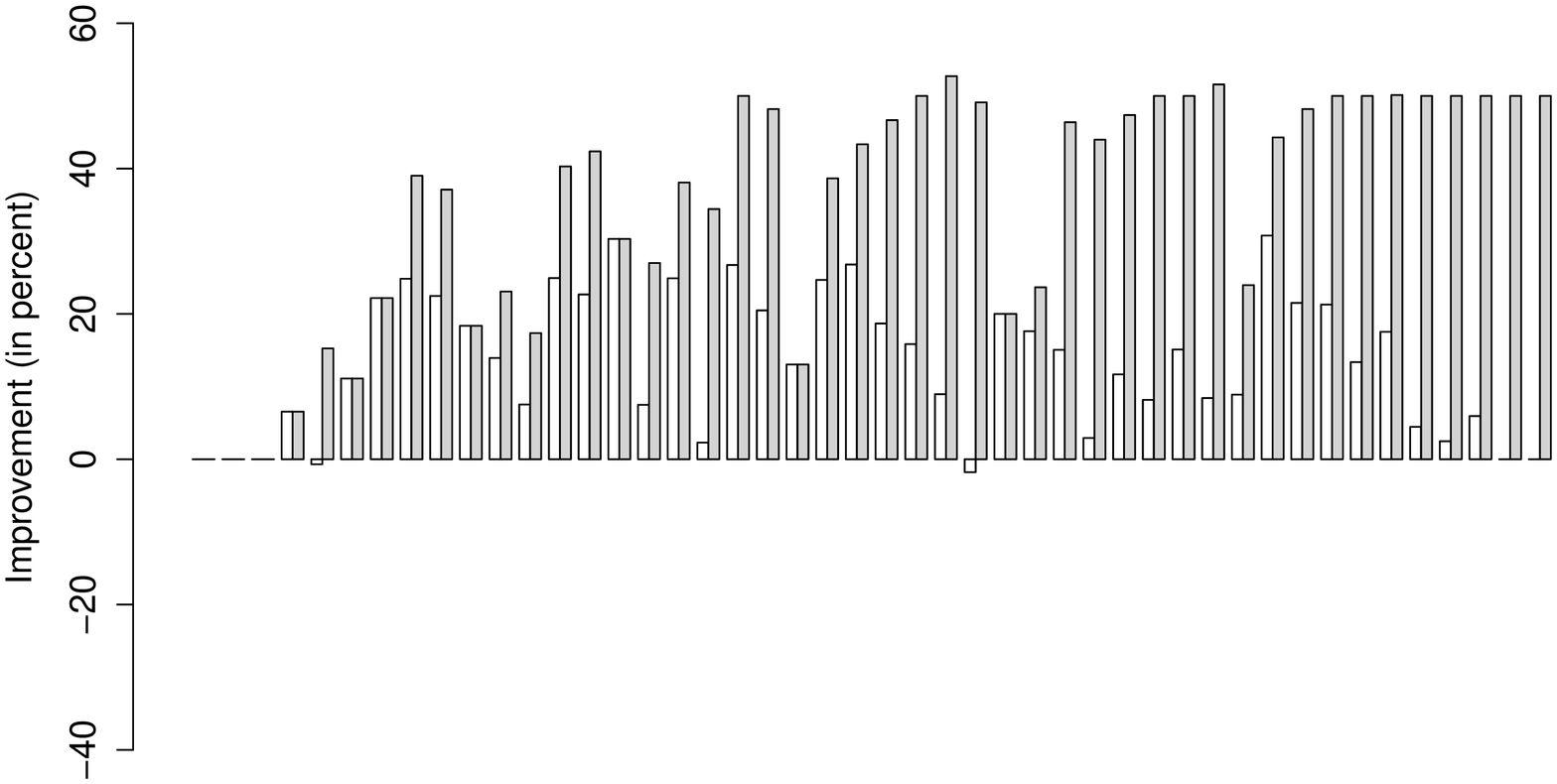}
\caption{Summary of results for grid and torus topologies. Both graphics show the performance improvement of \fsim\ (light gray bars) and \fsimp\ (dark gray bars) over \fin\ (in percent). The instances of Table~\ref{tab:res-grid} are treated from left to right in the same order. The top graphic concerns the best colorings found, whereas the bottom graphic concerns the average solution quality.}
\label{fig:res-grid}
\end{figure}

Figure~\ref{fig:res-grid} summarizes graphically the results from Table~\ref{tab:res-grid}. Note that the $y$-axis is differently scaled than the other summarizing figures in this section due to plotted data requirements. The significant improvement of \fsimp\ over both \fin\ and \fsim\ can be nicely appreciated in these graphics. Also the growing advantage of the FrogSim algorithms over \fin\ can be seen by the fact that the height of the bars generally increases from left to right. Considering the bottom graphic, which concerns the average solutions quality, we can note that \fsim\ is much less robust than \fsimp.

\subsection{Results for Small Instances from~\cite{lee2008firefly}}

Finally, we present results obtained by the three algorithms for rather small instances used by S.~Lee in~\cite{lee2008firefly} for the evaluation of a firefly-inspired distributed graph coloring algorithm. We do not directly compare with the results presented in~\cite{lee2008firefly}, because the algorithm proposed in~\cite{lee2008firefly} assumes that the number of colors required for the coloring is known a priori, that is, the algorithm must be run for a pre-fixed number of colors. When graphs are large and chromatic numbers are unknown, such an algorithm is not practical. Anyway, FrogSim and the algorithm from~\cite{lee2008firefly} behave very similarly for most instances, with some exceptions: for \emph{hexagon}-based instances, \fsimp\ is not quite able to match the average results obtained by the algorithm from~\cite{lee2008firefly}. Moreover, concerning \emph{icosahedron.col}, the best solution by \fsimp\ is uses one color more than the best one by Lee's algorithm. On the other side, concerning \emph{4-partite-4-diff-sizes.col} and \emph{dodecahedron.col}, \fsimp\ improves over the average results obtained by Lee's algorithm.

\begin{table*}[htbp]
\caption{Results of the algorithms on instances from the article~\cite{lee2008firefly}.}\label{tab:res-other}
\begin{center}
\scalebox{0.7}{
\begin{tabular}{ccccrrcrrcrr}
\hline \hline
\multirow{2}{*}{Instance} & $\;\;$ & \multirow{2}{*}{$(n,\Delta,\chi)$} & $\;\;$ & \multicolumn{2}{c}{Finocchi}& $\;\;$ & \multicolumn{2}{c}{$\text{FrogSim}$} & $\;\;$ &\multicolumn{2}{c}{$\text{FrogSim}^{\varoplus}$}\\
&& && colors & avg. && colors & avg. && colors & avg. \\
\cline{1-1} \cline{3-3} \cline{5-6} \cline{8-9} \cline{11-12}
1hexagon-tess.col&&(7,6,3) && {\bf 3}  & 3.000 && 3 & 3.564 && 3 & 3.198\\
2-partite-size6.col&&(12,6,2) && {\bf 2}  & 2.000 && {\bf 2}  & 2.000 && {\bf 2}  & 2.000\\
2hexagon-tess.col&&(10,6,3) && 4 & 4.000 && 3 & 3.604 && {\bf 3}  & 3.337\\
3-partite-3-diff-sizes.col&&(6,5,3) && {\bf 3}  & 3.000 && {\bf 3}  & 3.000 && {\bf 3}  & 3.000\\
3-partite-size-6.col&&(18,12,3) && {\bf 3}  & 3.000 && {\bf 3}  & 3.000 && {\bf 3}  & 3.000\\
3hexagon-tess.col&&(12,6,3) && 4 & 5.120 && 3 & 3.554 && {\bf 3}  & 3.307\\
3partite6.col&&(18,12,3) && {\bf 3}  & 3.000 && {\bf 3}  & 3.000 && {\bf 3}  & 3.000\\
4-partite-4-diff-sizes.col&&(10,9,4) && {\bf 4}  & 4.000 && {\bf 4}  & 4.000 && {\bf 4}  & 4.000\\
4triangles&&(12,3,?) && 3 & 3.770 && 3 & 3.149 && {\bf 3}  & 3.000\\
6hexagon-tess.col&&(19,6,3) && 5 & 5.000 && 4 & 4.634 && {\bf 4}  & 4.257\\
7partite2.col&&(14,12,7) && {\bf 7}  & 7.000 && {\bf 7}  & 7.000 && {\bf 7}  & 7.000\\
dodecahedron.col&&(20,3,3) && 3 & 3.570 && {\bf 3}  & 3.000 && {\bf 3}  & 3.000\\
icosahedron.col&&(12,5,4) && 5 & 5.000 && {\bf 4}  & 4.257 && {\bf 4}  & 4.257\\
peterson.col&&(10,3,3) && {\bf 3}  & 3.000 && {\bf 3}  & 3.000 && {\bf 3}  & 3.000\\
\cline{1-1} \cline{3-3} \cline{5-6} \cline{8-9} \cline{11-12}
average && && 3.714 & 3.890 && 3.429 & 3.626 && 3.429 & 3.525\\
\cline{1-1} \cline{3-3} \cline{5-6} \cline{8-9} \cline{11-12}
\# times better && && \multicolumn{2}{c}{1} && \multicolumn{2}{c}{0} && \multicolumn{2}{c}{4}\\
\# times all equal && &&\multicolumn{2}{c}{7} && \multicolumn{2}{c}{7} && \multicolumn{2}{c}{7}\\
\# times worse && &&\multicolumn{2}{c}{6} && \multicolumn{2}{c}{1} && \multicolumn{2}{c}{0}\\
\hline \hline
\end{tabular}}
\end{center}
\end{table*}

As shown in Table~\ref{tab:res-other}, the three algorithms achieve equal results in 7 out of 14 cases. Only in one case (see 1hexagon-tess.col) \fin\ is slightly better than the FrogSim algorithms due to the fact that it achieves an optimal coloring in all 100 applications. In the remaining cases both \fsim\ and \fsimp\ obtain better results than \fin. Moreover, it is remarkable that both \fsim\ and \fsimp\ obtain for 12 of the 14 instances optimal solutions. The difference between \fsim\ and \fsimp\ is again to be found in the fact that \fsimp\ is more robust, which is indicated by a better average solution quality.

\begin{figure}[ht!]
\centering
 \includegraphics[width=5cm,angle=-90]{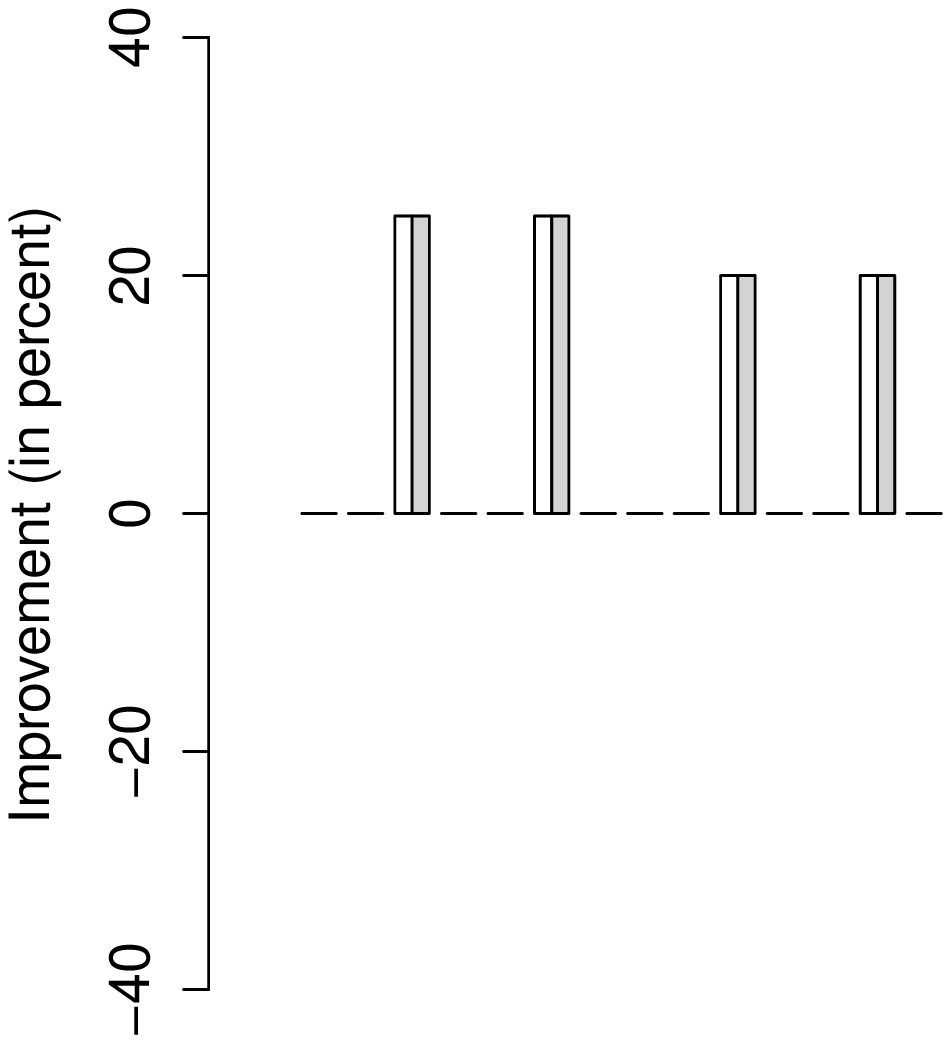}
 \includegraphics[width=5cm,angle=-90]{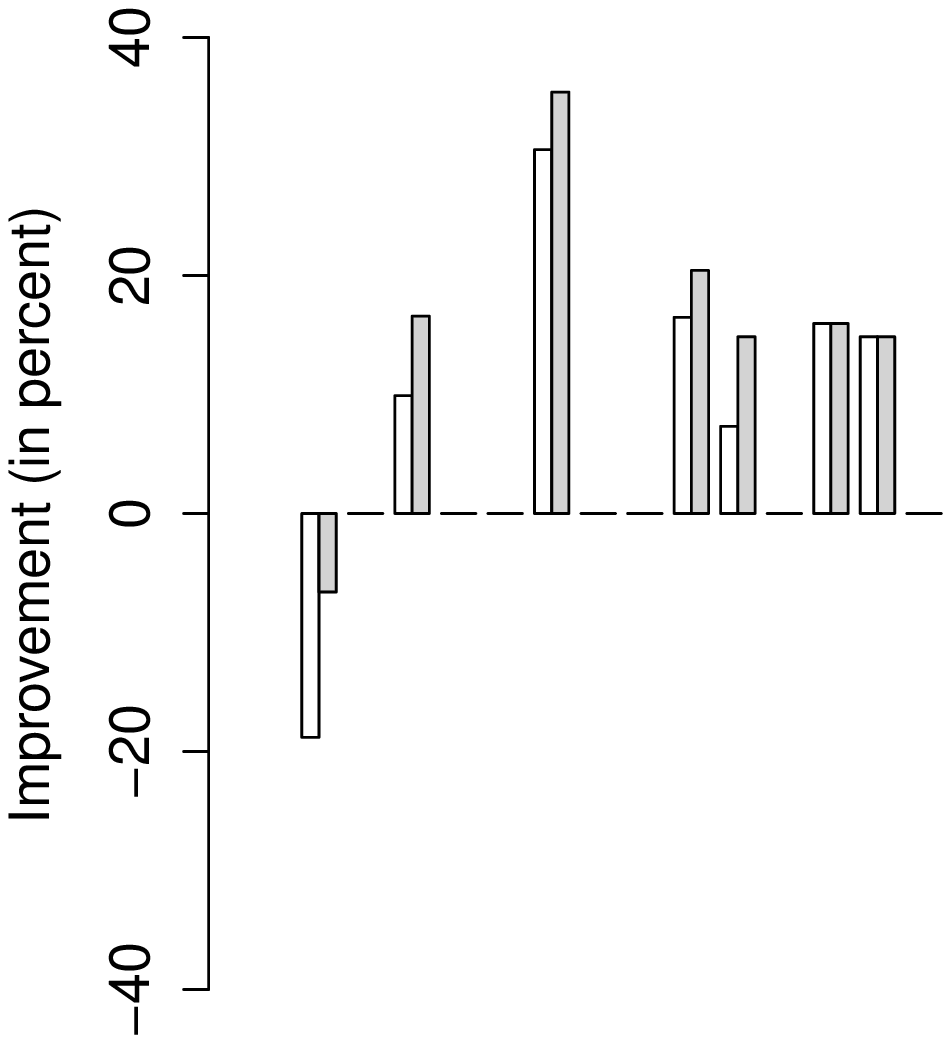}
\caption{Summary of results for the small graphs from~\cite{lee2008firefly}. Both graphics show the performance improvement of \fsim\ (light gray bars) and \fsimp\ (dark gray bars) over \fin\ (in percent). The instances of Table~\ref{tab:res-grid} are treated from left to right in the same order. The top graphic concerns the best colorings found, whereas the bottom graphic concerns the average solution quality.}
\label{fig:res-other}
\end{figure}

Figure~\ref{fig:res-other} graphically summarizes the results as in the previous subsections. Again this graphical way of presenting the results helps to show the improvement of \fsimp\ over \fsim\ in terms of the average solution quality. 

\section{Conclusions and Future Work}
\label{sec:conclusions}

Graph coloring is a classical problem of modern mathematics with more than $150$ years of history. The problem has been extensively studied in theory and practice. However, its connection to problems that have arisen with the proliferation of wireless networks has sparked a special interest in resolving the problem in a distributed manner. In such algorithms---due to the lack of global knowledge---the nodes have to base their color choices exclusively on information they receive from their direct neighborhood.

The algorithm we have presented in this paper is inspired by the behavior of a family of frogs native to Japan, namely the calling behavior of Japanese tree frogs. The results achieved by the proposed algorithm compare very favorably with current state-of-the-art algorithms. In particular, an improved performance has been measured for about $90\%$ of the studied instances. The benchmark set that we chose for comparison includes random geometric graphs, most of the graphs of the DIMACS challenge, and grid graphs. Apart from the favorable results, the proposed algorithms comes with some other benefits. It is possible, for example, to adjust the speed of convergence depending on the time the user wants to spend on the algorithm. Moreover, the number of communication rounds required is comparable to that required by other algorithms that provide high quality solutions. Finally, our algorithm provides a valid coloring already in the very first communication round. 

With regard to future work, we consider the use of the proposed algorithm for \emph{time division multiplexing} (TDM) which is a mechanism for collision-free communication in wireless networks, which is strongly related to graph coloring. Finally, due to its adaptive nature, our algorithm might also be interesting for mobile networks, or any dynamically changing network. The fact that nodes appear or disappear at certain points in time is nothing strange in wireless ad hoc networks.

\section*{Acknowledgment}

This work was supported by grant TIN2007-66523 (FORMALISM) of the Spanish government, and by the EU project FRONTS (FP7-ICT-2007-1). In addition, C.~Blum acknowledges support from the \textit{Ram{\'o}n y Cajal} program of the Spanish Government, and H.~Hern{\'a}ndez acknowledges support from the \emph{Comissionat per a Universitats i Recerca del Departament d'Innovaci{\'o}, Universitats i Empresa de la Generalitat de Catalunya} and from the \emph{European Social Fund}.

%
% ---- Bibliography ----
%
%\bibliographystyle{plain}
%\bibliography{graph-coloring}

\end{document}